\documentclass[accepted]{uai2021} 

\usepackage{enumerate}
\usepackage{enumitem}
\usepackage[T1]{fontenc}
\usepackage{inputenc}
\usepackage{verbatim}

\usepackage{tgtermes}
\usepackage{bm,amsbsy}
\usepackage{amsmath,amsfonts,amsthm,amssymb}
\usepackage[noend]{algpseudocode}
\usepackage{scalefnt,letltxmacro}
\LetLtxMacro{\oldtextsc}{\textsc}
\renewcommand{\textsc}[1]{\oldtextsc{\scalefont{1.10}#1}}
\usepackage[scaled=0.92]{PTSans}

\usepackage{cuted}

\usepackage[usenames,dvipsnames]{xcolor}
\definecolor{shadecolor}{gray}{0.9}

\usepackage[colorinlistoftodos,
           prependcaption,
           textsize=small,
           backgroundcolor=yellow,
           linecolor=lightgray,
           bordercolor=lightgray
           ]{todonotes}

\usepackage{lineno}

\usepackage{ragged2e}

\DeclareRobustCommand{\parhead}[1]{\textbf{#1}~}

\usepackage{graphicx}
\usepackage[labelfont=bf]{caption}
\usepackage{subfigure}
\usepackage{pstricks, pst-node}

\usepackage{booktabs}
\usepackage{multirow}
\usepackage{float}

\usepackage{paralist}
\usepackage{enumitem} 

\usepackage[algoruled,linesnumbered]{algorithm2e}
\usepackage{listings}
\usepackage{fancyvrb}
\fvset{fontsize=\normalsize}

\usepackage[all]{hypcap}
\hypersetup{colorlinks}
\hypersetup{linktoc=all}
\hypersetup{citecolor=Violet}
\hypersetup{linkcolor=black}
\hypersetup{urlcolor=MidnightBlue}

\theoremstyle{plain}

\theoremstyle{plain}
\newtheorem{assumption}{\protect\assumptionname}
\theoremstyle{plain}
\newtheorem{prop}{\protect\propositionname}
\theoremstyle{plain}
\newtheorem{lem}{\protect\lemmaname}

\providecommand{\assumptionname}{Assumption}
\providecommand{\lemmaname}{Lemma}
\providecommand{\propositionname}{Proposition}
\providecommand{\theoremname}{Theorem}

\usepackage[nameinlink]{cleveref}
\creflabelformat{equation}{#1#2#3}
\Crefname{equation}{Eq.}{Eqs.}
\Crefname{assumption}{Assumption}{Assumptions}
\Crefname{prop}{Proposition}{Propositions}
\Crefname{lem}{Lemma}{Lemmas}

\usepackage[acronym,nowarn]{glossaries}
\usepackage{listings}
\lstdefinestyle{alp_style}{
    commentstyle=\color{OliveGreen},
    numberstyle=\tiny\color{black!60},
    stringstyle=\color{BrickRed},
    basicstyle=\ttfamily\scriptsize,
    breakatwhitespace=false,
    breaklines=true,
    captionpos=b,
    keepspaces=true,
    numbers=none,
    numbersep=5pt,
    showspaces=false,
    showstringspaces=false,
    showtabs=false,
    tabsize=2
}
\usepackage{xr}
\makeatletter
\newcommand*{\addFileDependency}[1]{
  \typeout{(#1)}
  \@addtofilelist{#1}
  \IfFileExists{#1}{}{\typeout{No file #1.}}
}
\makeatother

\DeclareRobustCommand{\E}[2]{\mathbb{E}_{#1}\left[#2\right]}

\DeclareRobustCommand{\trace}[1]{\textrm{Tr}}

\newcommand{\g}{\, | \,}

\newcommand{\Ncal}{\mathcal{N}}

\newcommand{\Reals}{\mathbb{R}}

\newcommand{\cK}{\mathcal{K}}
\newcommand{\cL}{\mathcal{L}}
\newcommand{\cU}{\mathcal{U}}
\newcommand{\cZ}{\mathcal{Z}}

\newcommand{\loglik}{\cL}
\newcommand{\elbo}{\cL_{\text{ELBO}}}
\newcommand{\iwae}{\cL_{\text{IWAE}}}
\newcommand{\obs}{x}

\newcommand{\lat}{z}
\newcommand{\cLat}{\cZ}

\newcommand{\kernel}{\cK}
\newcommand{\decp}{\theta}
\newcommand{\encp}{\phi}
\newcommand{\coupl}[1]{\bar{#1}}
\newcommand{\indic}{\ell}
\newcommand{\corr}{\beta}
\newacronym{AIS}{AIS}{annealed importance sampling}
\newacronym{DISIR}{DISIR}{dependent iterated sampling importance resampling}
\newacronym{ELBO}{ELBO}{evidence lower bound}
\newacronym{ESS}{ESS}{effective sample size}
\newacronym{HMC}{HMC}{Hamiltonian Monte Carlo}
\newacronym{ISIR}{ISIR}{iterated sampling importance resampling}
\newacronym{IWAE}{IWAE}{importance weighted auto-encoder}
\newacronym{KL}{KL}{Kullback-Leibler}
\newacronym{MCMC}{MCMC}{Markov chain Monte Carlo}
\newacronym{MH}{MH}{Metropolis-Hastings}
\newacronym{ML}{ML}{maximum likelihood}
\newacronym{MLP}{MLP}{feed forward neural network}
\newacronym{PIMH}{PIMH}{particle independent Metropolis-Hastings}
\newacronym{PPCA}{PPCA}{probabilistic principal component analysis}
\newacronym{SMC}{SMC}{sequential Monte Carlo}
\newacronym{VAE}{VAE}{variational auto-encoder}
\newacronym{VI}{vi}{variational inference}

\usepackage[american]{babel}
\usepackage{natbib} %
\usepackage{mathtools} %
\usepackage{booktabs} %
\usepackage{tikz} %

\newcommand{\comm}[1]{}

\title{Unbiased Gradient Estimation for Variational Auto-Encoders\\using Coupled Markov Chains}

\author[1]{Francisco J.\ R.\ Ruiz} %
\author{Michalis K.\ Titsias}
\author{Taylan Cemgil}
\author{Arnaud Doucet}
\affil{
    DeepMind
}
\begin{document}
\maketitle

\begin{abstract}
    The \gls{VAE} is a deep latent variable model that has two neural networks in an autoencoder-like architecture; one of them parameterizes the model's likelihood. Fitting its parameters via \gls{ML} is challenging since the computation of the marginal likelihood involves an intractable integral over the latent space; thus the \gls{VAE} is trained instead by maximizing a variational lower bound. Here, we develop a \gls{ML} training scheme for \glspl{VAE} by introducing unbiased estimators of the log-likelihood gradient. We obtain the estimators by augmenting the latent space with a set of importance samples, similarly to the \gls{IWAE}, and then constructing a \acrlong{MCMC} coupling procedure on this augmented space. We provide the conditions under which the estimators can be computed in finite time and with finite variance. We show experimentally that \glspl{VAE} fitted with unbiased estimators exhibit better predictive performance.
\end{abstract}

\glsresetall

\section{INTRODUCTION}
\label{sec:introduction}

The \gls{VAE} \citep{Kingma2014} is a deep latent variable model that uses a joint distribution $p_{\decp}(\obs,\lat)$, parameterized by $\decp$, over an observation $\obs$ and the corresponding latent variable $\lat$. The marginal log-likelihood involves an integral over the latent space,
\begin{equation}\label{eq:log-marginal likelihood}
    \loglik(\decp):= \log p_{\decp}(\obs)=\log \left(\int p_{\decp}(\obs,\lat) \textrm{d}\lat\right).
\end{equation}
As for any other latent variable model, fitting the \gls{VAE} requires finding the parameters $\decp$ that best describe the observations. One (intractable) way to find $\decp$ would be via maximum likelihood, for which the gradient of \Cref{eq:log-marginal likelihood} is required. Using Fisher's identity, this gradient can be written as an expectation w.r.t.\ the posterior $p_{\decp}(\lat \g \obs)$, %
\begin{equation}\label{eq:scoreFisheridentity}
    \nabla_{\decp} \loglik(\theta)
    =
    \E{p_{\decp}(\lat \g \obs)}{ \nabla_{\decp}\log  p_{\decp}(\obs,\lat)}.
\end{equation}
The gradient in \Cref{eq:scoreFisheridentity} could be approximated unbiasedly if we had access to samples from $p_{\decp}(\lat \g \obs)$; however, the posterior is intractable. Although we could use \gls{MCMC} to sample approximately from it \citep{hoffman2017learning,naesseth2020markovian}, this would provide a biased estimate, and the bias is difficult to quantify.

Instead, \glspl{VAE} introduce an encoder $q_{\encp}(\lat\g\obs)$ and learn the parameters $\decp$ by maximizing a variational \gls{ELBO} \citep{Wainwright2008,Blei2017}, for which unbiased gradients are readily available \citep{Kingma2014}. 
\citet{Burda2016} form such a bound using a set of importance samples using the encoder as proposal distribution, leading to the so-called \gls{IWAE}. The standard \gls{ELBO} in variational inference can be thought of as a particular instance of the \gls{IWAE} bound with one importance sample, where the importance distribution is given by the encoder. However, it remains difficult to quantify the difference between the true log-likelihood and the corresponding bound.

We develop here unbiased gradient estimators of the log-likelihood for \glspl{VAE} by exploiting the coupling estimators developed by \cite{jacob2017unbiased}. 
Coupling estimators allow us to obtain unbiased estimators of expectations w.r.t.\ an intractable target distribution by running two coupled \gls{MCMC} chains for a finite number of iterations. This approach does not require that the \gls{MCMC} chains converge to the target, so the unbiased estimator can be computed in a finite (but random) time. However, \gls{MCMC} couplings are a generic methodology that is not readily applicable to \glspl{VAE}, since it requires an \gls{MCMC} kernel that mixes well and provides a suitable coupling mechanism, while at the same time yielding a low-variance estimator.

We address these issues by building coupling estimators based on the \gls{ISIR} algorithm \citep{Andrieu2010}. \gls{ISIR} is in turn based on importance sampling, and thus it uses multiple samples to reduce the variance of the estimator, similarly to the \gls{IWAE}.
Like the \gls{IWAE}, we simultaneously fit an encoder as the proposal distribution for the importance sampling algorithm.
Unfortunately, the variance of \gls{ISIR} is still not small enough for fitting \glspl{VAE}. To that end, we develop an extension of \gls{ISIR}, called \gls{DISIR}, that combines the use of dependent importance samples and reparameterization ideas. \gls{DISIR} drastically reduces the running time and the variance of the resulting coupling estimator. We develop a \gls{DISIR}-based coupling estimator to estimate the gradient of the \gls{VAE} log-likelihood.

\parhead{Contributions.}
Our contributions are as follows.
\begin{compactitem}
    \item We show that the unbiased gradient estimators based on \gls{ISIR} are not practical for \glspl{VAE}, since they require to run the Markov chains for a long time even for only moderately high-dimensional models. 
    \item We develop \gls{DISIR}, an extension of \gls{ISIR} that reduces the running time and the variance of the estimators.
    \item We use \gls{DISIR} to form unbiased gradient estimators for \glspl{VAE}. The resulting estimator is widely applicable as it can be used wherever the \gls{IWAE} bound is applicable.
    \item We prove that the unbiased gradient estimates have finite variance and can be computed in finite expected time under some regularity conditions.
    \item We demonstrate experimentally that \glspl{VAE} fitted with the \gls{DISIR}-based gradient estimators exhibit better predictive log-likelihood on binarized MNIST, fashion-MNIST, and CIFAR-10, when compared to \gls{IWAE}.
\end{compactitem}
\section{BACKGROUND}
\label{sec:background}

Here, we review the \gls{IWAE} bound \citep{Burda2016} and show that it can be seen as a standard \gls{ELBO} on an augmented model.
Consider a model $p_{\decp}(\obs,\lat)$ of data $x$ and latent variables $z\in \cZ$, and a proposal distribution $q_{\encp}(\lat \g \obs)$, where $\decp$ and $\encp$ denote the model and proposal parameters, respectively. Let the proposal satisfy the assumption below.\looseness=-1
\begin{assumption}[The proposal and posterior have the same support]
    \label{ass:supportok}
    For any $\lat\in\cLat$, we have $q_{\encp}(\lat \g \obs)>0$ if and only if $p_{\decp}(\obs, \lat)>0$, so that
    $0 < w_{\decp,\encp}(z) < \infty$, where the \emph{importance weights} are
    \begin{equation}\label{eq:weight_function}
        w_{\decp,\encp}(z) := \frac{p_{\decp}(\obs, \lat)}{q_{\encp}(\lat \g \obs)}.
    \end{equation}
\end{assumption}

\parhead{The \gls{IWAE} bound.}
The \gls{IWAE} is a lower bound of the marginal log-likelihood in \Cref{eq:log-marginal likelihood} formed with $K\geq 1$ importance samples $\lat_{1:K}$ from the proposal, i.e., $\loglik(\decp) \geq \iwae(\decp, \encp)$, with
\begin{equation}\label{eq:iwae}
    \iwae(\decp, \encp) =  \mathbb{E}_{q_{\encp}(\lat_{1:K} \g \obs)}\!\!\left[\log\! \left(\!\frac{1}{K} \sum_{k=1}^{K} w_{\decp,\encp}(\lat_k)\!\right)\!\right]\!,
\end{equation}
where $q_{\encp}(\lat_{1:K} \g \obs) := \prod_{k=1}^K q_{\encp}(\lat_k \g \obs)$.
\Cref{eq:iwae} monotonically increases with $K$, converging towards 
$\loglik(\decp)$
as $K\rightarrow\infty$.
For the case $K=1$, it recovers the standard \gls{ELBO},
\begin{equation}\label{eq:elbo}
    \elbo(\decp, \encp) =  \E{q_{\encp}(\lat \g \obs)}{\log w_{\decp,\encp}(\lat)}.
\end{equation}
The importance samples $\lat_{1:K}$
also provide an approximation of the posterior $p_{\decp}(\lat \g \obs)$. Specifically, if we define the importance weights $w_{\decp,\encp}^{(k)} := w_{\decp,\encp}(\lat_k)$ and the normalized importance weights $\widetilde{w}_{\decp,\encp}^{(k)} \propto  w_{\decp,\encp}^{(k)}$, with $\sum_{k=1}^K \widetilde{w}_{\decp,\encp}^{(k)}=1$, then the approximation is
\begin{equation}
    \hat{p}_{\decp}(\lat \g \obs)=\sum_{k=1}^K \widetilde{w}_{\decp,\encp}^{(k)} \delta_{\lat_k}(\lat),
    \label{eq:iwae_approx_posterior}
\end{equation}
where $\delta_{\lat_k}(\cdot)$ is the delta Dirac measure located at $\lat_k$.

\parhead{Fitting \glspl{VAE}.}
\glspl{VAE} parameterize the likelihood $p_{\decp}(\obs \g \lat )$ using a distribution whose parameters are given by a neural network (decoder) that inputs the latent variable $\lat$. The distribution $q_{\encp}(\lat \g \obs)$ is amortized \citep{gershman2014amortized}, i.e., its parameters are computed by a neural network (encoder) that inputs the observation $\obs$. Fitting a \gls{VAE} involves maximizing the bound (either \Cref{eq:iwae} or \Cref{eq:elbo}) w.r.t.\ both $\decp$ and $\encp$ using stochastic optimization. For that, the \gls{VAE} uses unbiased gradient estimators of the objective. To form such gradients, we typically assume that $q_{\encp}(\lat \g \obs)$ is reparameterizable \citep{Kingma2014,Rezende2014,Titsias2014_doubly}. For convenience, we additionally assume that we can reparameterize in terms of a Gaussian (but we can easily relax this latter assumption).
\begin{assumption}[The variational distribution is reparameterizable in terms of a Gaussian]
    \label{ass:reparamtrick}
    There exists a mapping $g_{\encp}(\xi, \obs)$ such that by sampling $\xi \sim q(\xi)$, where $q(\xi)=\mathcal{N}(\xi;0,I)$, and setting $\lat = g_{\encp}(\xi, \obs)$, we obtain $\lat \sim q_{\encp}(\lat \g \obs)$.
\end{assumption}

\parhead{The \gls{IWAE} bound as a standard \gls{ELBO}.}
The \gls{IWAE} bound in \Cref{eq:iwae} can be interpreted as a regular \gls{ELBO} on an augmented latent space \citep{Cremer2017,Domke2018}, and we use this perspective in \Cref{sec:method,sec:couplingMCMC}.
Indeed, consider the $K$ importance samples $\lat_{1:K}$ and an indicator variable $\indic\in\{1, \ldots, K\}$. We next define a generative model with latent variables $(\lat_{1:K}, \indic)$, as well as a variational distribution, such that its \gls{ELBO} recovers \Cref{eq:iwae}.

The augmented generative model posits that the indicator $\indic \sim \textrm{Cat}(\frac{1}{K},\ldots,\frac{1}{K})$,
where $\text{Cat}$ denotes the categorical distribution. Given $\indic$, each $\lat_k$ is distributed according to $q_{\encp}(\lat \g \obs)$, except the $\indic$-th one, which follows the prior: %
\begin{equation}\label{eq:p_augmented_joint}
    p_{\decp,\encp}(\obs, \lat_{1:K},\indic) = 
    \frac{1}{K} \; p_{\decp}(\obs, \lat_{\indic}) \prod_{k=1,k\neq \indic}^{K} q_{\encp}(\lat_k \g \obs).
\end{equation}
Under the corresponding augmented posterior distribution, $p_{\decp,\encp}(\lat_{1:K},\indic \g \obs)\propto  p_{\decp,\encp}(\obs, \lat_{1:K},\indic)$, the random variable $\lat_{\indic}$ follows the posterior $p_{\decp}(\lat \g \obs)$. We next define a variational distribution on the same augmented space,
\begin{equation}\label{eq:q_augmented_post}
    q_{\decp,\encp}(\lat_{1:K},\indic \g \obs) = 
    \text{Cat}\! \left(\!\indic \g \widetilde{w}_{\decp,\encp}^{(1)},\cdots,\widetilde{w}_{\decp,\encp}^{(K)} \!\right)
    \!\prod_{k=1}^{K} q_{\encp}(\lat_k \g \obs).
\end{equation}
We recover the \gls{IWAE} bound (\Cref{eq:iwae}) as the \gls{ELBO} of the augmented model, i.e., as $\E{q_{\decp,\encp}(\lat_{1:K},\indic \g \obs)}{\log p_{\decp,\encp}(\obs, \lat_{1:K},\indic)}$.

\parhead{Towards unbiased gradient estimation.}
The gradient w.r.t.\ $\decp$ of the \gls{IWAE} bound in \Cref{eq:iwae} can be interpreted as a (biased) approximation of %
$\nabla_{\theta}\loglik$ from \Cref{eq:scoreFisheridentity}. To see this, %
note that $\nabla_{\decp}\iwae
= \E{\hat{p}_{\decp}(\lat \g \obs)}{ \nabla_{\decp}\log  p_{\decp}(\obs,\lat)}$; i.e., $\nabla_{\decp}\iwae$ can be seen as an approximation of $\nabla_{\decp} \loglik$ where we replace the posterior $p_{\decp}(\lat \g \obs)$ with the approximation $\hat{p}_{\decp}(\lat \g \obs)$ in \Cref{eq:iwae_approx_posterior}.

To obtain an unbiased estimate, we need an alternative approximation $\hat{p}_{\decp}(\lat \g \obs)$ of $p_{\decp}(\lat \g \obs)$ satisfying 
$\E{\hat{p}_{\decp}(\lat \g \obs)}{
 \nabla_{\decp} \log p_{\decp}(\obs,\lat)} =\nabla_{\decp} \loglik$.
One such example is the empirical measure of exact samples from $p_{\decp}(\lat \g \obs)$. However, as mentioned earlier, it is typically impossible to obtain such samples, as a finite run with an \gls{MCMC} kernel only provides biased estimates when the chain is initialized out of equilibrium. We can obtain an approximation $\hat{p}_{\decp}(\lat \g \obs)$ that leads to unbiased estimation using two coupled \gls{MCMC} chains
\citep{jacob2017unbiased}. %
In \Cref{sec:method,sec:couplingMCMC} we build on this idea to develop a method for unbiased gradient estimation.\looseness=-1

\section{IMPORTANCE SAMPLING-BASED MCMC SCHEMES}
\label{sec:method}

\subsection{Iterated Sampling Importance Resampling (\acrshort{ISIR})}
\label{subsec:coupledISIR}

Here we review \gls{ISIR} \citep{Andrieu2010}, an \gls{MCMC} scheme that samples from the augmented posterior $p_{\decp,\encp}(\lat_{1:K}, \indic \g \obs)$ given by \Cref{eq:p_augmented_joint}. Alternatively, \gls{ISIR} can be interpreted as an algorithm that targets the posterior $p_{\decp}(\lat\g \obs)$. Indeed, if $(\lat_{1:K}, \indic) \sim p_{\decp,\encp}(\lat_{1:K}, \indic \g \obs)$ is a sample from the augmented posterior, then $\lat_{\indic}\sim p_{\decp}(\lat \g \obs)$.

The \gls{ISIR} transition kernel, $\kernel_{\textrm{ISIR}}(\cdot, \cdot  \g \lat_{1:K}, \indic)$, takes the current state $(\lat_{1:K}, \indic)$ and outputs a new state by following \Cref{alg:ISIR}. \gls{ISIR} requires a proposal distribution; we use $q_{\encp}(\lat \g \obs)$. It also requires to compute the importance weights $w_{\decp,\encp}(\cdot)$ in \Cref{eq:weight_function}.
Note that \gls{ISIR} has a resampling step (\Cref{alg_step:isir_sample_lstar}) that is distinct from the resampling step of sequential Monte Carlo, which would involve resampling multiple times from the categorical before mutating the samples.\looseness=-1

The kernel is invariant w.r.t.\ $p_{\decp,\encp}(\lat_{1:K}, \indic\g \obs)$, as formalized in \Cref{prop:ISIR} in \Cref{app:subsec:proof_boundISIR}, if \Cref{ass:weightsbounded} below is satisfied \citep{Andrieu2010,andrieu2018uniform}. \Cref{ass:weightsbounded} holds when the proposal is at least as heavy-tailed as the target.

\begin{assumption}[The importance weights are bounded]
    \label{ass:weightsbounded}
    There exists $w_{\decp,\encp}^{\textrm{max}} < \infty$ such that $w_{\decp,\encp}(\lat) \leq w_{\decp,\encp}^{\textrm{max}}$ $\forall \lat \in \cLat$.
\end{assumption}

\subsection{Dependent Iterated Sampling Importance Resampling (\acrshort{DISIR})}
\label{subsec:correlated_noise}

\begin{figure}[t]
    \centering
    \subfigure[Target density, proposal, and weight function.]{\centering \includegraphics[width=0.45\textwidth]{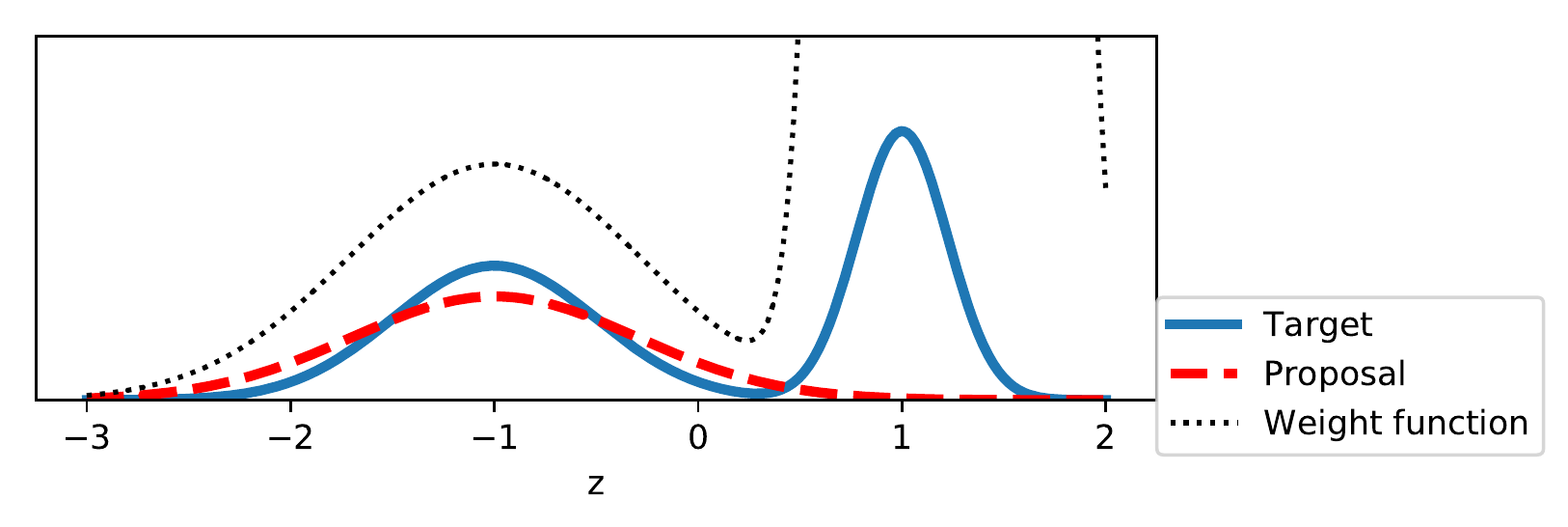}}\\ \vspace*{-5pt}
    \subfigure[A realization from the \acrshort{ISIR} kernel.]{\centering\includegraphics[width=0.45\textwidth]{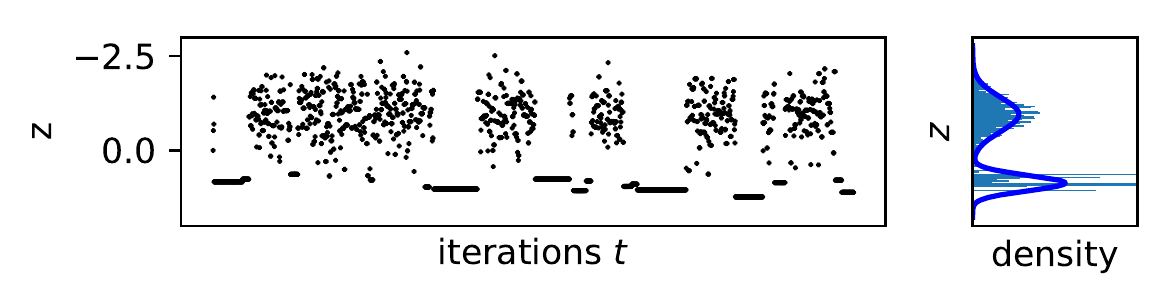}}\\ \vspace*{-5pt}
    \subfigure[A realization from the \acrshort{DISIR} kernel.]{\centering\includegraphics[width=0.45\textwidth]{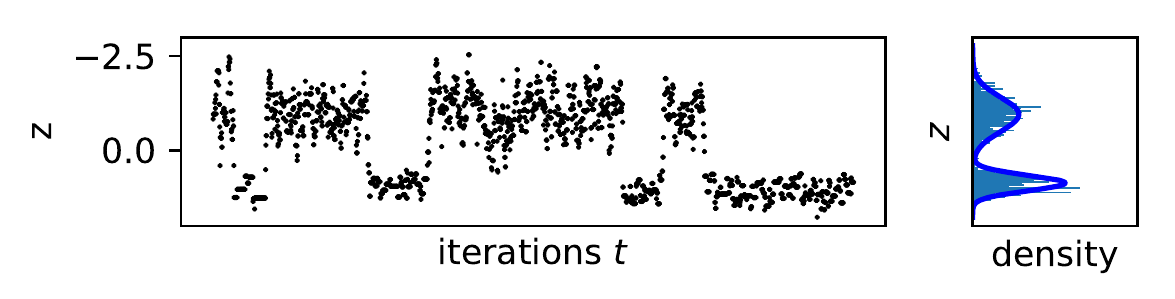}}
    \vspace*{-5pt}
    \caption{Qualitative comparison of \acrshort{ISIR} and \acrshort{DISIR} targetting a simple target distribution (a), and the realized chains by sampling from \acrshort{ISIR} (b) and \acrshort{DISIR} (c) transition kernels. Due to the poor proposal choice, the weight function significantly varies across the space $\lat$. \acrshort{ISIR} has a low acceptance probability, especially in high-weight states. In contrast, \acrshort{DISIR} is able to propose and accept local moves around high-weight regions and explores the target better.%
    \label{fig:illustrative_example}}
\end{figure}

\begin{algorithm}[t]
    \SetAlgoLined
    \DontPrintSemicolon
    \caption{%
        \acrshort{ISIR} kernel, $\kernel_{\textrm{ISIR}}(\cdot, \cdot \g \lat_{1:K}, \indic)$ %
        \label{alg:ISIR}
    }
    \SetKwComment{mycomment}{$\triangleright$ }{}
    \KwIn{Current state of the chain, $(\lat_{1:K}, \indic)$}
    \KwOut{Next state of the chain}
    Sample $\indic_{\textrm{aux}} \sim \textrm{Cat}(\frac{1}{K},\ldots,\frac{1}{K})$\;
    Set $\lat^{\star}_{\indic_{\textrm{aux}}} = \lat_{\indic}$ \;
    Sample $\lat_k^{\star} \sim q_{\encp}(\lat \g \obs)$ for $k\in\{1,\ldots, K\}\backslash \{\indic_{\textrm{aux}}\}$ \label{alg_step:isir_sample_zk} \;
    Sample $\indic^{\star} \sim \textrm{Cat}(p_1,\ldots,p_K)$ with $p_k\propto w_{\decp,\encp}(\lat_k^\star)$
    \label{alg_step:isir_sample_lstar}\;
    Return $(\lat_{1:K}^{\star}, \indic^{\star})$\;
\end{algorithm}

For moderately high-dimensional $\lat$, \gls{ISIR} can be inefficient. Indeed, if the importance weights become dominated by the weight of a single sample, then the corresponding Markov chain will typically get ``stuck'' for a large number of iterations. 
We illustrate this in \Cref{fig:illustrative_example}, which shows a one-dimensional illustrative comparison. In this toy experiment, we can observe that the \acrshort{ISIR} chains gets stuck when the state $\lat^{(t)}$ corresponds to a high-weight region. %

To mitigate this problem, here we develop \gls{DISIR}, an extension of \gls{ISIR} that uses dependent importance samples. Intuitively, this scheme proposes dependent samples $\lat_{1:\indic_{\textrm{aux}}-1}^{\star},\lat_{\indic_{\textrm{aux}}+1:K}^{\star}$ that are close to the current sample $\lat_{\indic_{\textrm{aux}}}^\star=\lat_{\indic}$ with high probability, in the spirit of \citet{shestopaloff2018sampling}. This modification increases the probability that the chain transitions to one of the new proposed values.
As a result, in practice, the gradient estimators based on \gls{DISIR} have smaller variance than the ones based on \gls{ISIR}.

To make samples dependent, we use the reparameterization property (see \Cref{ass:reparamtrick}) and introduce dependencies among the auxiliary variables $\xi_{1:K}$; this induces dependencies among the samples in the original latent space $\lat_{1:K}$. That is, rather than sampling $K-1$ importance samples  ($\xi^\star_{1:\indic_{\textrm{aux}}-1},\xi^\star_{\indic_{\textrm{aux}}+1:K}$) independently of $\xi^\star_{\indic_{\textrm{aux}}}$ (see \Cref{alg_step:isir_sample_zk} of \Cref{alg:ISIR}), we use two auxiliary Markov chains, each with transition kernel $p_{\corr}(\xi^\star \g \xi)$. Specifically, given $\xi^\star_{\indic_{\textrm{aux}}}$, we sample $\xi_{k}^\star \sim p_{\corr}(\cdot \g \xi_{k-1}^\star)$ for $k>\indic_{\textrm{aux}}$ and $\xi_{k}^\star \sim p_{\corr}(\cdot \g \xi_{k+1}^\star)$ for $k<\indic_{\textrm{aux}}$.
The kernel $p_{\corr}(\xi^\star \g \xi)$ must have invariant density $q(\xi)$, so that marginally $\xi_{k}^\star \sim  q(\xi)$ for all $k$ if $\xi^\star_{\indic_{\textrm{aux}}} \sim q(\xi)$. This construction with two Markov chains ensures the validity of the scheme (see \Cref{app:subsec:proof_invarianceDISIR} for details).

Under \Cref{ass:reparamtrick}, we select a simple autoregressive normal kernel for $p_{\corr}(\xi^\star \g \xi)$, i.e., we set
\begin{equation}
    \label{eq:kernel_corr_noise}
    p_{\corr}(\xi^\star \g \xi)=\mathcal{N}(\xi^\star; \corr \xi, (1-\corr^2) I).
\end{equation}
Equivalently, $\xi^\star =\corr \xi + \sqrt{1-\corr^2} \xi^{\textrm{new}}$, where $\xi^{\textrm{new}}\sim \mathcal{N}(\xi;0,I)$. The parameter $\corr$ controls the strength of the correlation; we discuss its effect below.

Using the kernel in \Cref{eq:kernel_corr_noise}, we develop \gls{DISIR}, which is described in \Cref{alg:DISIR}.
\gls{DISIR} replaces \Cref{alg_step:isir_sample_zk} of \Cref{alg:ISIR} with an application of the auxiliary kernel $p_{\corr}(\xi^\star \g \xi)$ (see \Cref{alg_step:disir_xi1,alg_step:disir_xi2} of \Cref{alg:DISIR}).

We refer to the coefficient $\corr\in[0, 1)$ as the \emph{correlation strength}. When $\corr=0$, we have $\xi_k^\star = \xi_k^{\textrm{new}} \overset{\textrm{iid}}{\sim} q(\xi)$ and this approach is simply a reparameterized version of \gls{ISIR}, which favours exploration of new regions of the space.
When $\corr$ approaches $1$, all the proposed values $\xi_k^\star$ become closer to the current state $\xi_{\indic}$ (which may correspond to the sample whose importance weight currently dominates). This dependency among the samples $\xi_{1:K}^\star$ induces dependencies among $\lat_{1:K}^\star$, resulting in more uniform importance weights. Thus, we say that this approach favours exploitation.

As given in \Cref{alg:DISIR}, \gls{DISIR} is an \gls{MCMC} scheme that targets the augmented density provided below.

\begin{prop}[Invariant distribution of \acrshort{DISIR}]
    \label{prop:DISIR}
    Let \Cref{ass:supportok,ass:reparamtrick} hold.
    For any $K\geq 2$ and any $\corr \in [0,1)$, the \gls{DISIR} transition kernel $\kernel_{\textrm{DISIR}}$ admits
    \begin{align}
        \label{eq:augmentedtargetcorrelated}
        p_{\decp,\encp}^{\text{\emph{DISIR}}}(\xi_{1:K},\indic \g \obs) &= \frac{1}{K} \frac{w_{\decp,\encp}( g_{\encp}(\xi_\indic,x))q(\xi_\indic)}{p_{\decp}(x)}\\
        &\times \prod_{k=1}^{\indic-1} p_\corr(\xi_{k} \g \xi_{k+1}) \prod_{k=\indic+1}^K p_\corr(\xi_{k} \g \xi_{k-1})\nonumber
    \end{align}
    as invariant distribution and is ergodic.
\end{prop}
The proofs of all propositions and lemmas are provided in \Cref{app:sec:proofs}.

\begin{algorithm}[tb]
    \SetAlgoLined
    \DontPrintSemicolon
    \caption{%
        \acrshort{DISIR} kernel, $\kernel_{\textrm{DISIR}}(\cdot, \cdot \g \xi_{1:K}, \indic)$ 
        \label{alg:DISIR}
    }
    \SetKwComment{mycomment}{$\triangleright$ }{}
    \KwIn{Current state of the chain $(\xi_{1:K}, \indic)$ and correlation strength $\corr$}
    \KwOut{New state of the chain}
    Sample $\indic_{\textrm{aux}} \sim \textrm{Cat}(\frac{1}{K},\ldots,\frac{1}{K})$\;

    Set $\xi_{\indic_{\textrm{aux}}}^\star = \xi_{\indic}$\;
    
    Sample $\xi_k^{\textrm{new}} \sim q(\xi)$ for $k\in\{1,\ldots, K\}\backslash \{\indic_{\textrm{aux}}\}$ \;

    Set $\xi_k^{\star} = \corr \xi_{k-1}^\star + \sqrt{1-\corr^2} \xi_k^{\textrm{new}}$ for $k=\indic_{\textrm{aux}}+1,\ldots,K$ \label{alg_step:disir_xi1}\;
    
    Set $\xi_k^{\star} = \corr \xi_{k+1}^\star +\sqrt{1-\corr^2} \xi_k^{\textrm{new}}$ for $k=\indic_{\textrm{aux}}-1,\ldots,1$ \label{alg_step:disir_xi2}\;
    
    Set
    $\lat_k^\star=g_{\encp}(\xi_k^\star, \obs)$  for $k=1,\ldots,K$\;
    
    Sample $\indic^{\star} \sim \textrm{Cat}(p_1,\ldots,p_K)$ with $p_k\propto w_{\decp,\encp}(\lat_k^\star)$\;
    
    Return $(\xi_{1:K}^\star, \indic^\star)$\;
\end{algorithm}

This target distribution has two desired properties. First, when the correlation strength $\corr \in [0,1)$, the $\indic$-th sample $\lat_\indic$ is distributed according to the posterior $p_{\decp}(\lat \g \obs)$.
That is, \gls{DISIR} defines a Markov chain that targets \Cref{eq:augmentedtargetcorrelated}, and $(\lat^{(t)}_{\indic^{(t)}})_{t\geq 0}$ is a Markov chain that converges to $p_{\decp}(\lat \g \obs)$. Second, when $\corr=0$, \gls{DISIR} becomes identical to a reparameterized version of \gls{ISIR}. That is, it simulates a Markov chain $(\xi_{1:K}^{(t)},\indic^{(t)})_{t\geq 0}$ such that, setting each $\lat_k^{(t)}=g_{\encp}(\xi_k^{(t)},\obs)$, the Markov chain $(\lat_{1:K}^{(t)},\indic^{(t)})_{t\geq 0}$ obeys a law that is identical to the one simulated by \gls{ISIR}. This is formalized below.

\begin{lem}[Distribution of \acrshort{DISIR} samples]
    \label{lemma:invariance}
    Let \Cref{ass:supportok,ass:reparamtrick} hold. For any $\corr \in [0,1)$, we have $\lat_{\indic}=g_{\encp}(\xi_\indic,\obs) \sim p_{\decp}(\lat \g \obs)$ under $p_{\decp,\encp}^{\text{\emph{DISIR}}}(\xi_{1:K},\indic \g \obs)$. Moreover, for $\corr=0$, if $(\xi_{1:K},\indic) \sim p_{\decp,\encp}^{\text{\emph{DISIR}}}(\xi_{1:K},\indic \g \obs)$, then we have $(\lat_{1:K},\indic) \sim p_{\decp,\encp}(\lat_{1:K},\indic \g \obs)$, where each $\lat_k=g_{\encp}(\xi_k,\obs)$.
\end{lem}

In practice, we interleave \gls{DISIR} steps for which $\corr > 0$ with steps for which $\corr=0$. Specifically, we define a composed kernel that consists of the consecutive application of two steps of \Cref{alg:DISIR}. The first step has $\corr=0$ and favours exploration; the second step has $\corr>0$ and favours exploitation. It is possible to interleave the two kernels since both can be reinterpreted as \gls{MCMC} kernels targeting $p_{\decp}(\lat \g \obs)$. We denote the composed kernel as $\kernel_{\textrm{ISIR-DISIR}}$.

\parhead{Choice of the correlation strength.} For the second step of the composed kernel, we wish to use a value $\corr$ close to $1$ to achieve exploitation, but not too close because then we will effectively have one importance sample repeated $K$ times. We set $\corr$ following a heuristic that is based on the \gls{ESS}, defined as $\textrm{ESS}=(\sum_{k=1}^K (\widetilde{w}_{\decp,\encp}^{(k)})^2)^{-1}$. As $\corr$ becomes closer to $1$, the \gls{ESS} becomes closer to $K$. We set the target \gls{ESS} to $0.3K$, and we update $\corr$ after each step of the kernel, so that the \gls{ESS} becomes closer to the target value. In particular, we apply the update rule $\corr \leftarrow \corr - 0.01 (\textrm{ESS} - 0.3K)$. We also constrain the resulting $\corr \in [10^{-6}, 1-10^{-6}]$ to avoid numerical issues.

Since we only modify $\corr$ between iterations of the algorithm (and not while running the algorithm), the invariance of the Markov kernel w.r.t.\ to the target still holds. This also implies that the estimators that we derive in \Cref{sec:couplingMCMC} are unbiased despite the adaptation of $\beta$.

\subsection{\acrshort{DISIR} Estimates of Expectations}
\label{subsec:estimates_expectations}

The \gls{DISIR} samples $(\lat^{(t)}_{\indic^{(t)}})_{t\geq 0}$ are distributed asymptotically as $p_{\decp}(\lat \g \obs)$ and, in the limit of infinite samples, we can estimate expectations under $p_{\decp}(\lat \g \obs)$, as $\tfrac{1}{T+1}\sum_{t=0}^{T} h(\lat^{(t)}_{\indic^{(t)}}) \rightarrow \mathbb{E}_{p_{\decp}(\lat \g \obs)}[h(z)]$ almost surely.

However, it may seem wasteful to generate $K-1$ proposals at each iteration of \Cref{alg:DISIR} and then use only the $\indic$-th importance sample to estimate an expectation. The proposition below shows that it is possible to use all the $K$ importance samples: we can estimate an expectation $\mathbb{E}_{p_{\decp}(\lat \g \obs)}[h(z)]$ with a weighted average of the importance samples.

\begin{prop}[The $K$ importance samples can be used for estimating expectations]
    \label{prop:identityexpectation}
    For any function $h:\cLat \rightarrow \mathbb{R}$ such that $\mathbb{E}_{p_{\decp}(\lat \g \obs)}[|h(\lat)|]< \infty$, we have the identity
    \begin{equation}\label{eq:identitiesconditionalexpectations}
        \mathbb{E}_{p_{\decp}(\lat \g \obs)}[h(z)]=\mathbb{E}_{p^{\text{\emph{DISIR}}}_{\decp,\encp}(\xi_{1:K},\indic \g \obs)}\!\!\left[\sum_{k=1}^K \widetilde{w}_{\decp,\encp}^{(k)} h(\lat_k)\right]\!,
    \end{equation}
    where $\lat_k = g_{\encp}(\xi_k,\obs)$ and the normalized importance weights are 
    $\widetilde{w}_{\decp,\encp}^{(k)} \propto w_{\decp,\encp}(\lat_k)$ with $\sum_{k=1}^K \widetilde{w}_{\decp,\encp}^{(k)}=1$.
    Setting $h(\lat)=\nabla_{\decp}\log p_{\decp} (x, \lat)$, and given that the \gls{DISIR} kernel is ergodic, it follows from \Cref{eq:scoreFisheridentity} that
    \begin{equation}
        \tfrac{1}{T+1}\sum_{t=0}^{T} \left[\sum_{k=1}^K \widetilde{w}_{\decp,\encp}^{(k,t)} \nabla _{\decp} \log p_{\decp}(\obs,\lat_k^{(t)})\right] \rightarrow 
        \nabla_{\decp} \loglik
    \end{equation}
    almost surely as $T \rightarrow \infty$ for any $K\geq 2$, where $\lat_k^{(t)}=g_{\encp}(\xi^{(t)}_k,\obs)$ and
    $\widetilde{w}_{\decp,\encp}^{(k,t)} \propto w_{\decp,\encp}(\lat_k^{(t)})$.
\end{prop}

\subsection{Choice of the Proposal}
\label{subsec:proposal_optimization}

Both \gls{ISIR} and \gls{DISIR} require a proposal distribution to sample the states $\lat_k^{\star}$, for which we use $q_{\encp}(\lat \g \obs)$ (\gls{DISIR} additionally requires the proposal to be reparameterizable). Here we discuss how to set the parameters $\encp$ of the proposal.

Like for \gls{IWAE}, in our case $q_{\encp}(\lat \g \obs)$ is a proposal distribution rather than a variational posterior.
We fit $\encp$ via stochastic optimization of the \gls{IWAE} bound in \Cref{eq:iwae}. To estimate the gradient w.r.t.\ $\encp$, we use the doubly reparameterized estimator \citep{Tucker2019}, which addresses some issues of the estimator of \citet{Burda2016} for large values of $K$ \citep{Rainforth2018}.

Alternatively, we could fit the encoder using the forward \gls{KL} divergence. This would imply to maximize $\E{p_{\decp}(\lat\g\obs)}{\log q_{\encp}(\lat\g\obs)}$ w.r.t.\ $\encp$, for which we can apply the unbiased estimators of \Cref{sec:couplingMCMC} to estimate the expectation w.r.t.\ $p_{\decp}(\lat\g\obs)$. We leave this for future work.
\section{UNBIASED GRADIENT ESTIMATION WITH MCMC COUPLINGS}
\label{sec:couplingMCMC}

\subsection{Unbiased Estimation with Couplings}
\label{subsec:unbiased_couplings}

In this section, we review how to obtain an unbiased gradient estimator using two coupled Markov chains. We use the notation $u \in \cU$ to refer to a generic random variable, keeping in mind that we will later set $u=[\xi_{1:K}, \indic]$, i.e., the latent variables in the augmented space.

Consider the estimation of the expectation
\begin{equation}\label{eq:expectation_h}
  H := \E{\pi(u)}{h(u)},
\end{equation}
for some distribution $\pi(u)$ and function $h(u)$.
As discussed in 
\Cref{sec:background},
a direct approximation $\hat{\pi}(u)$ of $\pi(u)$ via \gls{MCMC} leads to a biased estimator.

We can obtain an unbiased estimator based on two coupled \gls{MCMC} chains, each with invariant distribution $\pi(\cdot)$ \citep{glynn2014exact,jacob2017unbiased}. The two chains have the same marginals at any time instant $t$, but they evolve according to a joint transition kernel $\kernel_{\textrm{C}}$.
Let $\kernel(\cdot\g u)$ be the marginal transition kernel of each chain, and let $\kernel_{\textrm{C}}(\cdot, \cdot \g u, \coupl{u})$ be a joint kernel that takes the state of both chains (denoted $u$ and $\coupl{u}$) and produces the new state of both chains.\footnote{%
    The joint kernel is such that $\kernel_{\textrm{C}}(A, \cU \g u, \coupl{u}) = \kernel(A \g u)$ and $\kernel_{\textrm{C}}(\cU, A \g u, \coupl{u}) = \kernel(A \g \coupl{u})$ for any measurable set $A$.
}

We next review the unbiased estimator of \citet{VanettiDoucet2020} which reduces the variance of the estimator of \citet{jacob2017unbiased}. The main idea is to consider a lag $L\geq 1$ and jointly sample the states of both chains $(u^{(t)}, \coupl{u}^{(t-L)})$ conditioned on their previous states, i.e., $(u^{(t-1)}, \coupl{u}^{(t-L-1)})$.
We then use (a finite number of) the samples from each chain to obtain the unbiased estimator (\Cref{app:sec:review_couplings} shows how to derive it).
Practically, we initialize the first Markov chain from some (arbitrary) initial distribution $\pi_0(\cdot)$, i.e., $u^{(0)}\sim \pi_0(u)$. We then sample this Markov chain using the marginal kernel $\kernel$, i.e., $u^{(t)}\sim \kernel(u \g u^{(t-1)})$ for $t=1,\ldots,L$. After $L$ steps, we draw the initial state of the second Markov chain $\coupl{u}^{(0)}$ (potentially conditionally upon $u^{(L-1)},u^{(L)}$), such that marginally $\coupl{u}^{(0)}\sim \pi_0(u)$. Afterwards, for $t> L$, we draw both states jointly as $(u^{(t)}, \coupl{u}^{(t-L)}) \sim \kernel_{\textrm{C}}(u, \coupl{u} \g u^{(t-1)}, \coupl{u}^{(t-L-1)})$.

The joint kernel $\kernel_{\textrm{C}}$ is chosen such that, after some time, both chains produce the same exact realizations of the random variables, i.e., $u^{(t)}=\coupl{u}^{(t-L)}$ for $t\geq \tau$. Here, $\tau$ is the \emph{meeting time}, defined as the first time instant in which both chains meet,
$\tau=\inf\{t\geq L: u^{(t)}=\coupl{u}^{(t-L)}\}$ (it could be infinite, but we design the joint kernel so that $\tau$ is a random variable of finite expected value).

Based on this coupling procedure, the unbiased estimator of \Cref{eq:expectation_h} by \citet{VanettiDoucet2020} is (see \Cref{app:sec:review_couplings})
\begin{equation}
    \label{eq:estimator_vanetti}
    \hat{H} \! = \! \frac{1}{L} \! \left( \sum_{t=t_0}^{t_0 + L - 1} \!\!\!\!\! h(u^{(t)}) \! + \!\!\!\! \sum_{t=t_0+L}^{\tau-1} \!\!\!\! \left( \! h(u^{(t)}) \! - \! h(\coupl{u}^{(t-L)})\!\right)\!\!\right)\!\!,
\end{equation}
where $t_0$ is a constant that plays the role of the burn-in period, although it is not a burn-in period in the usual sense, since we do not require the Markov chains to converge. Indeed, the estimator in \Cref{eq:estimator_vanetti} requires us to run the coupled Markov chains until they meet each other. Given that we design the joint kernel such that the meeting time $\tau$ is finite, this implies that we obtain the unbiased estimator in finite time.

We can also think of this unbiased coupling procedure as providing an empirical approximation\footnote{%
    \Cref{eq:empirical_measure_vanetti} is a signed measure, i.e., we can have $\mathbb{E}_{\hat{\pi}(u)}[h(u)]<0$ even for a positive function $h(\cdot)$.
}
$\hat{\pi}(\cdot)$ of $\pi(\cdot)$,
\begin{equation}
    \label{eq:empirical_measure_vanetti}
    \hat{\pi}(u) \! = \! \frac{1}{L} \! \left( \sum_{t=t_0}^{t_0 + L - 1} \!\!\!\!\! \delta_{u^{(t)}}\!(u) \! + \!\!\!\! \sum_{t=t_0+L}^{\tau-1} \!\!\!\!\! \left( \delta_{u^{(t)}}\!(u) \! - \! \delta_{\coupl{u}^{(t-L)}}(u)\right) \!\! \right) \!\!.
\end{equation}

We next provide sufficient conditions that ensure that the estimator in \Cref{eq:estimator_vanetti} can be computed in expected finite time and has finite variance. These conditions are similar as for the original estimator of \citep{jacob2017unbiased,middleton2018unbiasedEJS} but the proof is slightly different.

\begin{prop}[The unbiased estimator can be computed in finite time and has finite variance]
    \label{prop:unbiasedcond}
    Assume the following conditions hold:
    \begin{compactenum}[a.]
        \item\label{item:convergence_markov_chain} (Convergence of the Markov chain.) Each of the two chains marginally starts from a distribution $\pi_0$, evolves according to a transition kernel $\kernel$ and is such that $\E{}{h(u^{(t)})}\rightarrow \E{\pi(u)}{h(u)}$ as $t\rightarrow\infty$.
        \item\label{item:finite_high_order_moment} (Finite high-order moment.) There exist $\eta>0$ and $D< \infty$ such that  $\E{}{\left|h(u^{(t)})\right|^{2+\eta}}\leq D$ \; $\forall t \geq 0$.
        \item\label{item:finite_tau} (Distribution of the meeting time.) There exists an almost surely finite meeting time  $\tau=\inf\{t\geq L: u^{(t)}=\coupl{u}^{(t-L)}\}$ such that $\mathbb{P}(\tau > t)\leq C t^{-\kappa}$ for some $C<\infty$ and $\kappa> 2(2 \eta^{-1}+1),$ where $\eta$ appears in Condition (\ref{item:finite_high_order_moment}).
        \item\label{item:chains_stay_together} (The chains stay together after meeting.) We have $u^{(t)} = \coupl{u}^{(t-L)}$ for all $t\geq \tau$.
    \end{compactenum}
    Then, \Cref{eq:estimator_vanetti} is an unbiased estimator of $\mathbb{E}_{\pi(u)}[h(u)]$ that can be computed in finite expected time and has finite variance.\looseness=-1
\end{prop}

Conditions (\ref{item:convergence_markov_chain}) and (\ref{item:chains_stay_together}) can be satisfied by careful design of the joint kernel $\kernel_{\textrm{C}}$. Condition (\ref{item:finite_high_order_moment}) is a mild integrability condition. Condition (\ref{item:finite_tau}) can be satisfied if the marginal kernel $\kernel$ is (only) polynomially ergodic and some additional mild irreducibility and aperiodicity conditions on the joint kernel $\kernel_{\textrm{C}}$ hold \citep{middleton2018unbiasedEJS}.

\subsection{Coupling \acrshort{DISIR}}
\label{subsec:coupledDISIR}

Here we describe the main algorithm of this paper: an estimator based on two coupled Markov chains, each evolving according to the \gls{DISIR} transition kernel from \Cref{subsec:correlated_noise}. That is, we build a joint kernel $\kernel_{\textrm{C}}$ for unbiased gradient estimation. We denote the joint kernel as $\kernel_{\textrm{C-DISIR}}((\cdot,\cdot), (\cdot,\cdot) \g (\xi_{1:K}, \indic), (\coupl{\xi}_{1:K}, \coupl{\indic}) )$. It inputs the current state of both Markov chains, $(\xi_{1:K},\indic)$ and $(\coupl{\xi}_{1:K}, \coupl{\indic})$, and returns their new states.

The coupled \gls{DISIR} kernel (C-\acrshort{DISIR}) is given in \Cref{alg:C-DISIR}. It resembles the \gls{DISIR} kernel of \Cref{alg:DISIR}, and in fact, as required, it behaves as $\kernel_{\textrm{DISIR}}(\cdot,\cdot \g \xi_{1:K},\indic)$ marginally if we ignore one of the two Markov chains. Thus, \Cref{alg:C-DISIR} guarantees that the marginal stationary distribution of each chain is $p^{\textrm{DISIR}}_{\decp,\encp}(\xi_{1:K}, \indic \g \obs)$ (\Cref{eq:augmentedtargetcorrelated}).

\begin{algorithm}[t]
    \SetAlgoLined
    \DontPrintSemicolon
    \caption{%
        C-\acrshort{DISIR} kernel for two coupled chains, $\kernel_{\textrm{C-DISIR}}((\cdot, \cdot), (\cdot, \cdot) \g (\xi_{1:K}, \indic), (\coupl{\xi}_{1:K}, \coupl{\indic}))$
        \label{alg:C-DISIR}
    }
    \SetKwComment{mycomment}{$\triangleright$ }{}
    \KwIn{Current state of both chains, $(\xi_{1:K}, \indic)$ and $(\coupl{\xi}_{1:K}, \coupl{\indic})$, and correlation strength $\corr$}
    \KwOut{New state of both chains}
    Sample $\indic_{\textrm{aux}} \sim \textrm{Cat}(\frac{1}{K},\ldots,\frac{1}{K})$ \;
    Set $\xi_{\indic_{\textrm{aux}}}^\star = \xi_{\indic}$
    and $\coupl{\xi}_{\indic_{\textrm{aux}}}^\star = \coupl{\xi}_{\coupl{\indic}}$ \;
    Sample $\xi_k^{\textrm{new}} \sim q(\xi)$ for $k\in\{1,\ldots, K\}\backslash \{\indic_{\textrm{aux}}\}$ \;
    Set $\xi_k^{\star} = \corr \xi_{k-1}^\star + \sqrt{1-\corr^2} \xi_k^{\textrm{new}}$ and
    $\coupl{\xi}_k^{\star} =\corr \coupl{\xi}_{k-1}^\star + \sqrt{1-\corr^2} \xi_k^{\textrm{new}} $
    for $k=\indic_{\textrm{aux}}+1,\ldots,K$\;
    Set $\xi_k^{\star} =\corr \xi_{k+1}^\star+ \sqrt{1-\corr^2} \xi_k^{\textrm{new}}$ and
    $\coupl{\xi}_k^{\star} = \corr \coupl{\xi}_{k+1}^\star+ \sqrt{1-\corr^2} \xi_k^{\textrm{new}}$
    for $k=\indic_{\textrm{aux}}-1,\ldots,1$ \;
    Set
    $\lat_k^\star=g_{\encp}(\xi_k^\star, \obs)$ and $\coupl{\lat}_k^\star=g_{\encp}(\coupl{\xi}_k^\star, \obs)$
    for $k=1,\ldots,K$\;
    Sample $\indic^\star, \coupl{\indic}^\star \sim \kernel_{\textrm{C-Cat}}(\indic, \coupl{\indic} \g (w_{\decp, \encp}(\lat_1^{\star}),\ldots, w_{\decp, \encp}(\lat_K^{\star})),$
    $(w_{\decp, \encp}(\coupl{\lat}_1^{\star}),\ldots, w_{\decp, \encp}(\coupl{\lat}_K^{\star})))$ from the maximal coupling kernel (\Cref{alg:C-Cat} in \Cref{app:sec:K_C_Cat}) \label{alg_step:coupled_disir_sample_indicators} \;
    Return $((\xi_{1:K}^\star, \indic^\star), (\coupl{\xi}_{1:K}^\star, \coupl{\indic}^\star))$
\end{algorithm}

The indicators $(\indic^\star, \coupl{\indic}^\star)$ are sampled jointly from a kernel $\kernel_{\textrm{C-Cat}}$ (\Cref{alg_step:coupled_disir_sample_indicators} of \Cref{alg:C-DISIR}), which is given in \Cref{app:sec:K_C_Cat}. This corresponds to the maximal coupling kernel\footnote{%
    A coupling procedure is maximal if it maximizes the probability that both chains meet.
}
for two categorical distributions \citep{lindvall2002lectures}.

When the correlation strength $\corr=0$, i.e., when \gls{DISIR} is equivalent to \gls{ISIR}, coupling may occur; that is, \Cref{alg:C-DISIR} may return the same state for both chains. To see this, note that when $\corr=0$, \Cref{alg:C-DISIR} shares the same values of the noise values generating the importance samples for both chains, i.e., $\xi_k^\star = \coupl{\xi}^\star_k$ for $k\neq \indic_{\textrm{aux}}$, while the $\indic_{\textrm{aux}}$-th importance sample is set to the current state of each chain, i.e., $\xi_{\indic_{\textrm{aux}}}^\star = \xi_{\indic}$ and $\coupl{\xi}_{\indic_{\textrm{aux}}}^\star = \coupl{\xi}_{\coupl{\indic}}$. Thus, if the indicators sampled in \Cref{alg_step:coupled_disir_sample_indicators} take the same value (i.e., $\indic^\star = \coupl{\indic}^\star$) and this value is different from $\indic_{\textrm{aux}}$, then both chains meet. After meeting, for any future iteration of the joint kernel, the states of both chains are guaranteed to be identical to each other. On the contrary, when the correlation strength $\corr \neq 0$, coupling cannot occur. However, for any $\corr \in [0, 1)$, \Cref{alg:C-DISIR} guarantees that the chains remain equal to each other once they have previously met.

We use a composed kernel that consists of the consecutive application of two steps of \Cref{alg:C-DISIR}. The first step has $\corr=0$; in this step the chains may meet each other. The second step has $\corr>0$, which favours exploitation. As discussed in \Cref{subsec:correlated_noise}, it is valid to combine these two kernels. We denote this composed joint kernel as  $\kernel_{\textrm{C-ISIR-DISIR}}$.

\begin{algorithm}[t]
    \SetAlgoLined
    \DontPrintSemicolon
    \caption{%
        Unbiased estimation with C-\acrshort{ISIR}-\acrshort{DISIR} %
        \label{alg:couplingISIR}
    }
    \KwIn{The constant $t_0$ and the lag $L$}
    \KwOut{An unbiased estimator of $\nabla_{\decp} \loglik$}
    \SetKwComment{mycomment}{$\triangleright$ }{}
    Initialize
    $\xi_k \sim q(\xi)$ and $\coupl{\xi}_k \sim q(\xi)$ for $k=1,\ldots, K$ \;
    Initialize $\indic\sim \textrm{Cat}(\frac{1}{K},\ldots,\frac{1}{K})$ and $\coupl{\indic}\sim \textrm{Cat}(\frac{1}{K},\ldots,\frac{1}{K})$ \;
    \For{$t=1,\ldots,L$}{
        Sample $(\xi_{1:K}^{(t)}, \indic^{(t)})\sim \kernel_{\textrm{ISIR-DISIR}}(\cdot, \cdot \g \xi_{1:K}^{(t-1)}, \indic^{(t-1)})$ (two steps of \Cref{alg:DISIR}, with $\beta=0$ then $\beta>0$) \;
    }
    Set the iteration $t=L$ \;
    \While{$t<t_0 + L - 1$ \emph{ or the two chains have not met}}{
        Sample $((\xi_{1:K}^{(t+1)}, \indic^{(t+1)}), (\coupl{\xi}_{1:K}^{(t-L+1)}, \coupl{\indic}^{(t-L+1)}))
        \sim \kernel_{\textrm{C-ISIR-DISIR}}((\cdot,\cdot), (\cdot, \cdot) \g (\xi_{1:K}^{(t)}, \indic^{(t)}), (\coupl{\xi}_{1:K}^{(t-L)}, \coupl{\indic}^{(t-L)}) )$
        (two steps of \Cref{alg:C-DISIR}, with $\beta=0$ then $\beta>0$)\;
        Increase $t\leftarrow t+1$\;
    }
    Return the estimator from \Cref{eq:estimator_vanetti} using the function $h(\xi_{1:K},\indic):=\sum_{k=1}^K \widetilde{w}_{\decp,\encp}^{(k)} \nabla_{\decp} \log p_{\decp}(\obs, \lat_k)$ \;
\end{algorithm}

\parhead{Unbiased gradient estimation with C-\acrshort{ISIR}-\acrshort{DISIR}.}
\Cref{alg:couplingISIR} describes the procedure that provides an unbiased estimator of $\nabla_{\decp} \loglik$. It samples two Markov chains, $(\xi_{1:K}^{(t)}, \indic^{(t)})$ and $(\coupl{\xi}_{1:K}^{(t)}, \coupl{\indic}^{(t)})$, by inducing a coupling between the state of the first chain at time $t$ and the state of the second chain at time $t-L$, where $L\geq 1$ is the lag. After both chains meet, it returns the unbiased gradient estimator using \Cref{eq:estimator_vanetti} for the function $h(\xi_{1:K}, \indic) := \sum_{k=1}^K \widetilde{w}_{\decp,\encp}^{(k)} \nabla_{\decp} \log p_{\decp}(\obs, \lat_k)$ (applying \Cref{prop:identityexpectation}), where $\lat_k = g_{\encp}(\xi_k,\obs)$ and the normalized importance weights are $\widetilde{w}_{\decp,\encp}^{(k)} \propto w_{\decp,\encp}(\lat_k)$. \Cref{alg:couplingISIR} provides a practical unbiased estimator, as we show next.\looseness=-1

\begin{prop}
    \label{prop:estimateok}
    Let \Cref{ass:supportok,ass:reparamtrick,ass:weightsbounded} hold and condition (\ref{item:finite_high_order_moment}) of \Cref{prop:unbiasedcond} be satisfied. For any $K\geq2$,
    \Cref{alg:couplingISIR} returns an unbiased estimator of $\nabla_{\decp} \loglik$ of finite variance that can be computed in finite expected time. Additionally, $\mathbb{E}[\tau]$ can be upper bounded by a quantity decreasing with $K$. %
\end{prop}

\section{RELATED WORK}
\label{sec:related}

Our estimator builds on previous work discussed in the former sections. We now review other related works.

\gls{ISIR}, as well as other particle \gls{MCMC} algorithms, has been previously used for smoothing in state-space models  \citep{Andrieu2010} and for (biased) estimation of $\nabla_{\decp} \loglik$ \citep{naesseth2020markovian}. Coupled variants of these algorithms have also been previously developed for unbiased smoothing \citep{jacob2019smoothing,middleton2019unbiased}. Indeed, \Cref{alg:C-DISIR} for $\corr=0$ has been used by \citet{jacob2019smoothing} (without reparameterization).
However, we found experimentally that the unbiased estimators based on coupled \gls{ISIR} suffer from high variance for moderately high dimensions, making them impractical for \glspl{VAE}. The estimators based on coupled \gls{DISIR} with $\corr\approx 1$ address this issue.

An unbiased estimator based on a coupled Gibbs sampler has also been presented for restricted Boltzmann machines \citep{qiu2019unbiased}, but this method is not applicable for \glspl{VAE}.
An alternative unbiased gradient estimator for \glspl{VAE}, based on Russian roulette ideas, was developed by \citet{luo2019sumo}. However, this estimator suffers from high variance (potentially infinite), and requires additional variance reduction methods such as gradient clipping, which defeats the purpose of unbiased gradient estimation. In our experiments, we use RMSProp and no gradient clipping is needed.

\citet{dieng2019reweighted} maximize the marginal log-likelihood of the data using an expectation maximization scheme that gives a consistent (but not unbiased) estimator. 

Finally, note that the coupling estimators approximate the gradient w.r.t.\ $\decp$, and are orthogonal to the methods that improve the expressiveness of the encoder $q_{\encp}(\lat\g\obs)$, such as
semi-implicit methods \citep{Yin2018,TitsiasRuiz2019} or normalizing flows \citep{Rezende2015,Kingma2016,Papamakarios2017,Tomczak2016,Tomczak2017,Dinh2017}, to name a few. These methods could be used together with the coupling estimators to obtain a more flexible proposal distribution, which could improve the mixing of the Markov chains.

\section{EXPERIMENTS}
\label{sec:experiments}

In \Cref{subsec:experiments_ppca}, we study the bias and variance of different estimators in an experiment where we have access to the exact gradient $\nabla_{\decp} \loglik$. In \Cref{subsec:experiments_vae}, we study the predictive performance of \glspl{VAE} trained with coupled \gls{DISIR} and show that models fitted with unbiased estimators outperform those fitted via \gls{ELBO} or \gls{IWAE} maximization.
We implement all the estimators in JAX \citep{jax2018github}.

\subsection{\Acrlong{PPCA}}
\label{subsec:experiments_ppca}

We first consider \gls{PPCA}, as for this model we have access to the exact gradient $\nabla_{\decp} \loglik$. The model is $p_{\decp}(\obs,\lat) = \Ncal(\lat; 0, I) \Ncal(\obs; \decp_0 + \decp_1^\top \lat, 0.1 I)$, where $\lat\in\Reals^{100}$. We randomly set the model parameters $\decp$ and fit a variational distribution $q_{\encp}(\lat \g \obs)$ by maximizing the \gls{IWAE} bound w.r.t.\ $\encp$ with $K=100$ importance samples on binarized MNIST \citep{Salakhutdinov2008_quantitative}. The distribution $q_{\encp}(\lat \g \obs)$ is a fully factorized Gaussian whose parameters depend linearly on $\obs$.

We obtain the exact gradient, $\nabla_{\decp}\loglik = \sum_{n=1}^N \nabla_{\decp}\log p_{\theta}(\obs_n)$, for a batch of $N=100$ datapoints, and we compare it against four gradient estimators. Two estimators are the gradients of the \gls{ELBO} and \gls{IWAE} bounds ($\hat{\nabla}_{\decp} \elbo$ and $\hat{\nabla}_{\decp} \iwae$). The third one is the unbiased estimator obtained with coupled \gls{ISIR}, i.e., a variant of \Cref{alg:couplingISIR} where we replace the $\kernel_{\textrm{C-ISIR-DISIR}}$ kernel with $\kernel_{\textrm{C-ISIR}}$ (which is equivalent to $\kernel_{\textrm{C-DISIR}}$ with correlation strength $\corr=0$). The fourth estimator is based on \Cref{alg:couplingISIR}. For all the estimators, we use the same (fixed) distribution $q_{\encp}(\lat \g \obs)$. For the coupling estimators, we set $t_0=1$ and lag $L=10$.\looseness=-1

\begin{figure}[t]
    \centering
    \includegraphics[width=0.8\columnwidth]{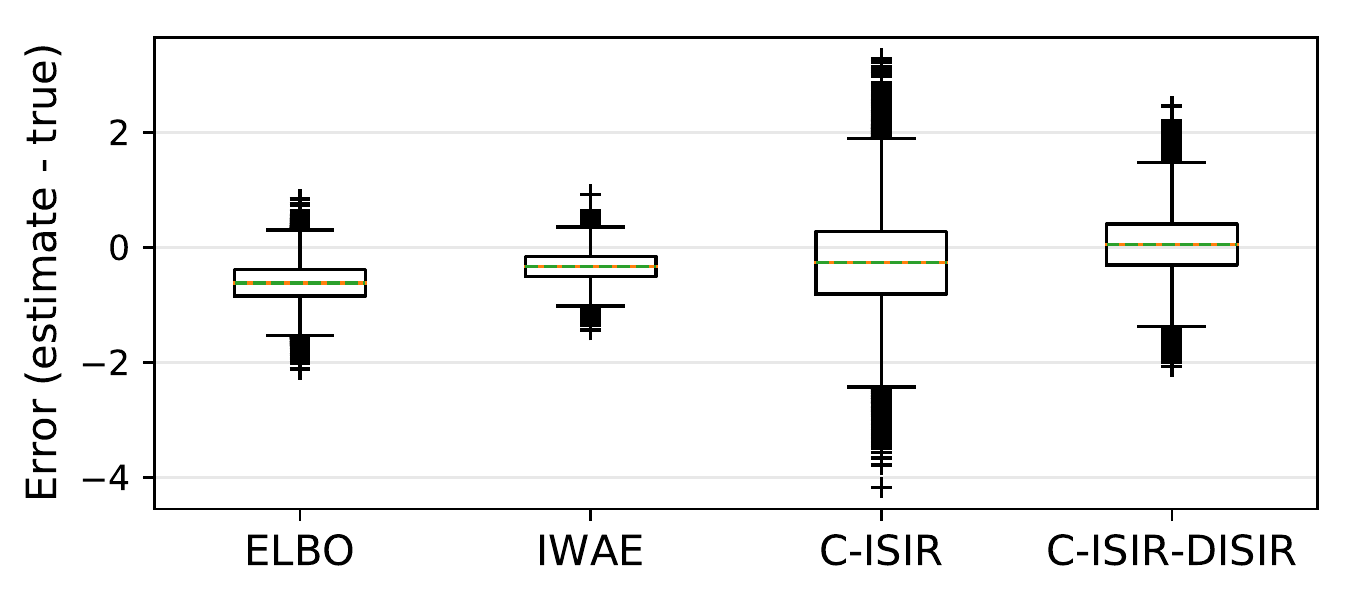}
    \vspace*{-8pt}
    \caption{Boxplot representation of the error of different estimators for the gradient w.r.t.\ one of the intercepts of the \acrshort{PPCA} model. The estimators based on variational bounds (\acrshort{ELBO} and \acrshort{IWAE}) are biased. Among the two unbiased estimators based on couplings, the one based on \Cref{alg:couplingISIR} (C-\acrshort{ISIR}-\acrshort{DISIR}) exhibits lower variance.%
    \label{fig:ppca_boxplots}}
    \vspace*{-8pt}
\end{figure}

We obtain $50{,}000$ samples from each estimator, and compute the (signed) error $\hat{\nabla}_{\decp} \loglik - \nabla_{\decp} \loglik$ for each sample. We show in \Cref{fig:ppca_boxplots} the boxplot representation of the error for a randomly chosen component of the gradient w.r.t.\ the intercept term. (In \Cref{app:subsec:ppca_grads}, we show a randomly chosen weight term in \Cref{fig:ppca_boxplots2}, and the average over components in \Cref{fig:ppca_boxplots_all}.) As expected, the estimators of the \gls{ELBO} and \gls{IWAE} gradients are biased. The boxplots for C-\acrshort{ISIR} and C-\acrshort{ISIR}-\acrshort{DISIR} are consistent with the unbiasedness of the estimators, and the one based on C-\acrshort{ISIR}-\acrshort{DISIR} has smaller variance. This property is key for fitting more complex models such as \glspl{VAE}.\looseness=-1

\subsection{\Acrlong{VAE}}
\label{subsec:experiments_vae}

Now we apply the coupling estimators to fit \glspl{VAE} and compare the predictive performance to the maximization of the \gls{ELBO} and \gls{IWAE} objectives. (We also implemented the method of \citet{luo2019sumo}, but we found it led to unstable optimization despite using gradient clipping.) We provide further details on the experimental setup in \Cref{app:subsec:vae_details}.

\parhead{Binarized MNIST.}
We first fit a \gls{VAE} on the statically binarized MNIST dataset. We use $K=10$ importance samples and explore the dimensionality $D=\{20, 100, 300\}$ of $\lat \in \Reals^D$. We use RMSProp \citep{Tieleman2012} in the stochastic optimization procedure. For the coupling estimators, we set $t_0=1$ step and the lag $L=10$.

Following \citet{wu2017quantitative}, we estimate the predictive log-likelihood using \gls{AIS} \citep{neal2001annealed}. Specifically, we use $16$ independent \gls{AIS} chains, with $10{,}000$ intermediate annealing distributions, and a transition operator consisting of one \gls{HMC} trajectory
with $10$ leapfrog steps and adaptive
acceptance rate tuned to
$0.65$. For CIFAR-10, we use $4$ \gls{AIS} chains with $7{,}500$ intermediate distributions and $5$ \gls{HMC} leapfrog steps.
As this procedure is computationally intensive, we only evaluate the train log-likelihood on the current data batch, but we evaluate the log-likelihood on the entire test set at the end of the optimization.

\begin{figure}[t]
    \centering
    \subfigure[Results on binarized MNIST for $D=100$. The unbiased estimator from \Cref{alg:couplingISIR} provides better performance.\label{fig:bmnist_loglik_D100}]{\includegraphics[width=\columnwidth]{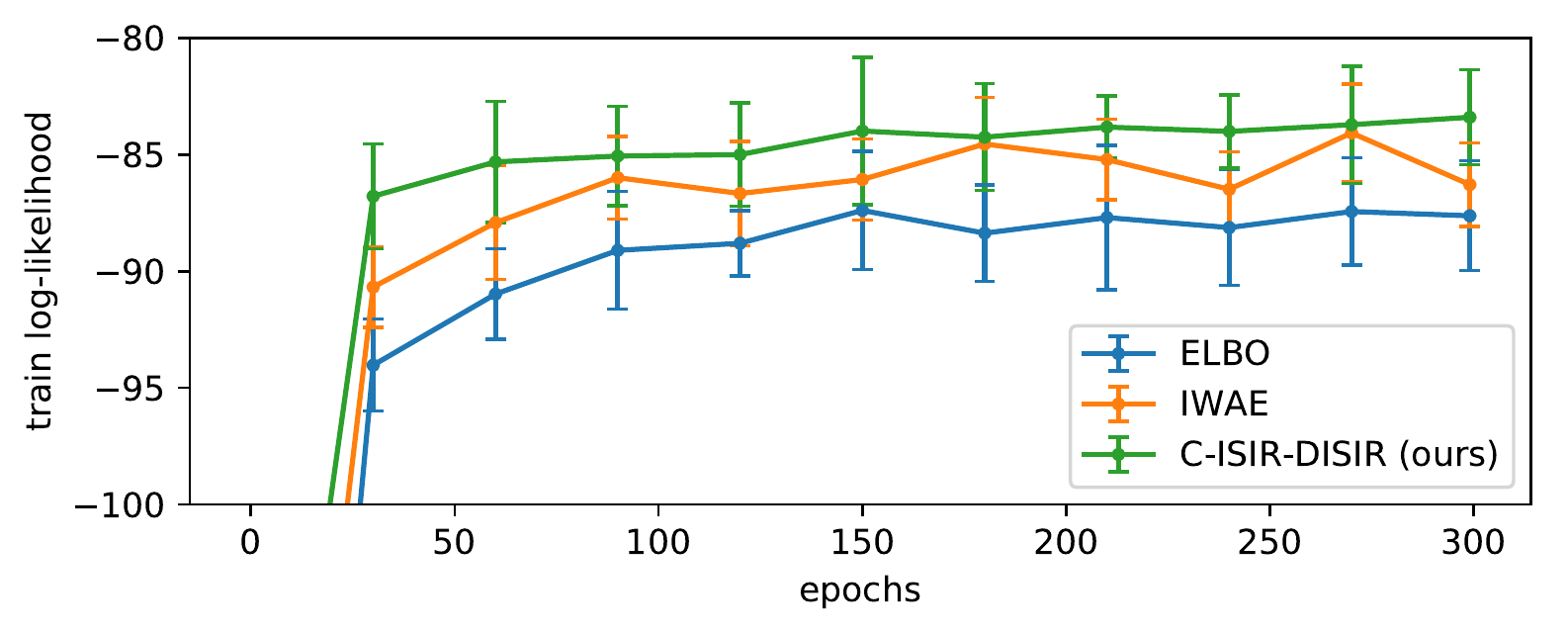}}\\ \vspace*{-8pt}
    \subfigure[Results on fashion-MNIST. After switching from the \gls{IWAE} to the unbiased estimator at epoch $200$, the performance improves.\label{fig:fashionmnist_loglik}]{\includegraphics[width=\columnwidth]{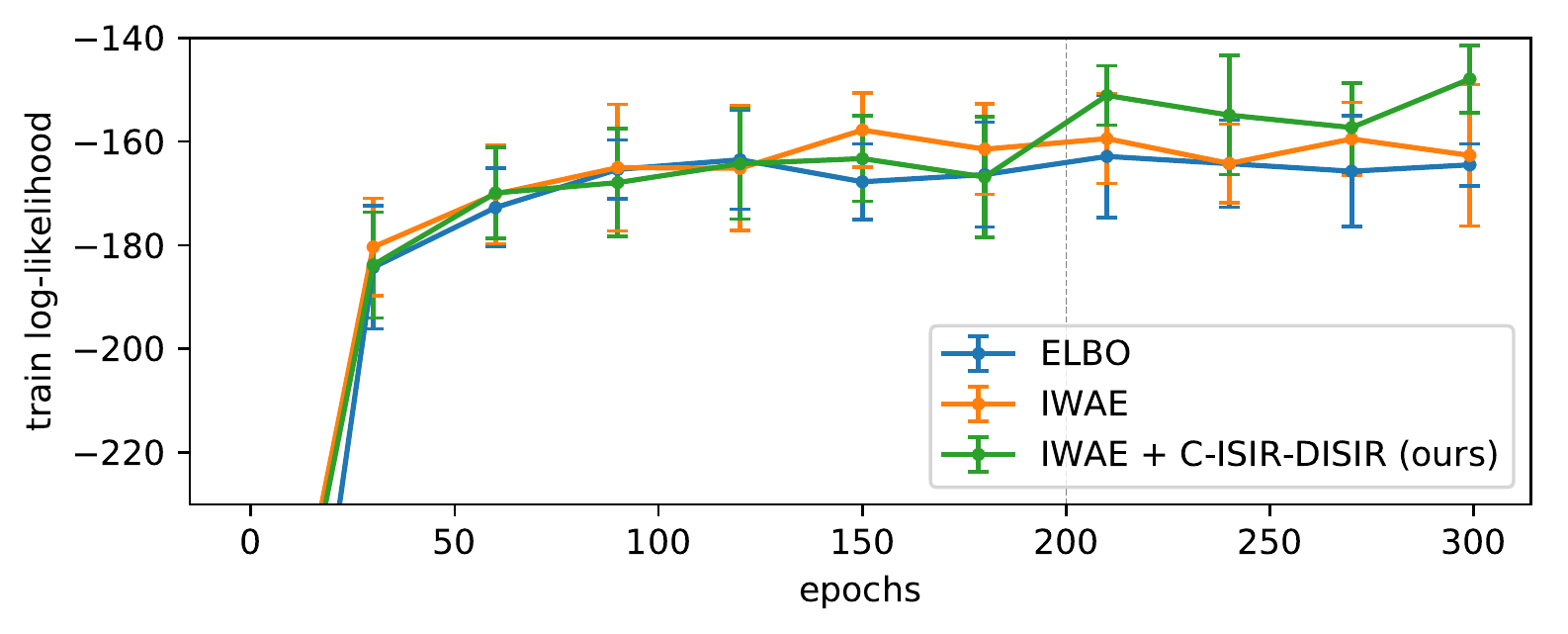}}  \vspace*{-4pt}
    \caption{Train log-likelihood for a \acrshort{VAE}.%
    \label{fig:vae_loglik}}
\end{figure}

\Cref{fig:bmnist_loglik_D100} shows the evolution of the train log-likelihood; the error bars correspond to the standard deviation of $10$ independent runs. (\Cref{fig:bmnist_loglik_all} in \Cref{app:subsec:train_log_lik} shows similar plots for varying $D$.) \Cref{tab:test_loglik_bmnist} shows the test log-likelihood after $300$ epochs. The \gls{VAE} models fitted with the unbiased estimator of \Cref{alg:couplingISIR} have better predictive performance.

\begin{figure}[ht]
    \centering
    \includegraphics[width=\columnwidth]{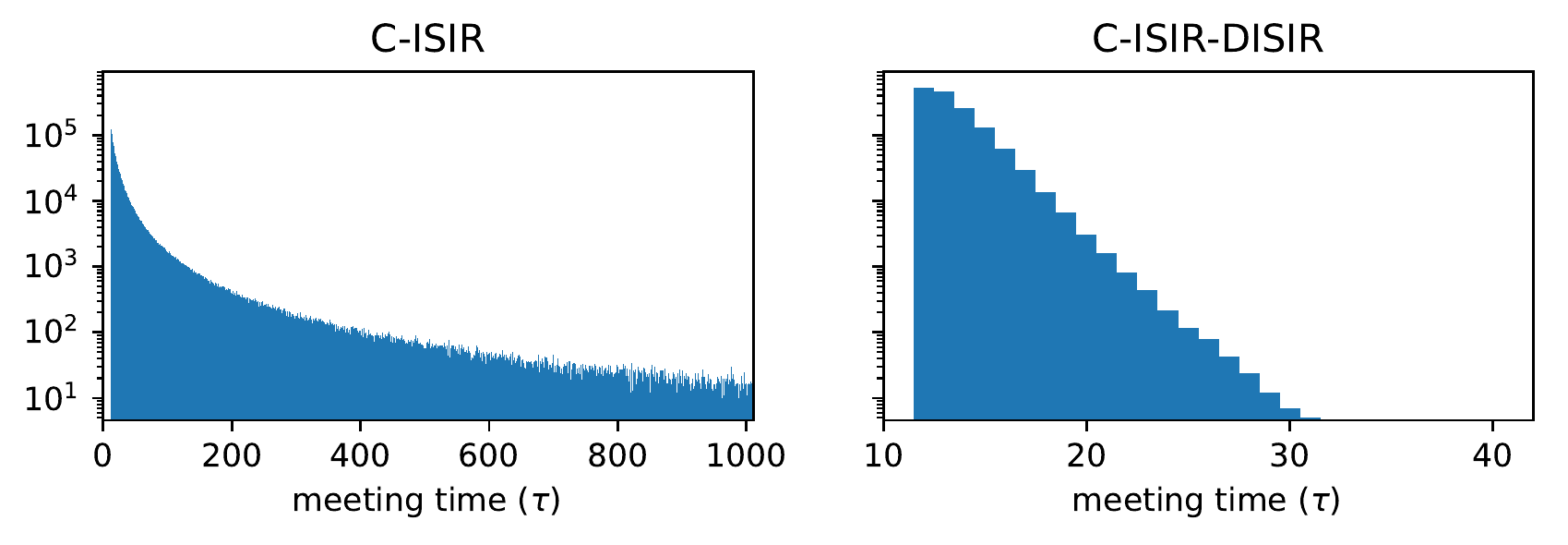}
    \vspace*{-18pt}
    \caption{Histogram of the meeting time for a \acrshort{VAE} fitted on binarized MNIST with $D=300$. The histogram corresponding to C-\acrshort{ISIR} has significantly heavier tails, which results in higher computational complexity of the overall estimator. Moreover, C-\acrshort{ISIR} occasionally ($1\%$) reaches the maximum allowed number of \acrshort{MCMC} iterations (hard-coded at around $1{,}000$), which induces a small bias in the estimator. C-\acrshort{ISIR}-\acrshort{DISIR} does not suffer from this issue.%
    \label{fig:bmnist_histograms_tau_D300}}
\end{figure}

The $\kernel_{\textrm{C-ISIR-DISIR}}$ kernel in \Cref{alg:couplingISIR} is key for obtaining this improved performance. As a comparison, replacing it with $\kernel_{\textrm{C-ISIR}}$ leads to a test log-likelihood value of $-90.70\pm 0.08$ for $D=20$, i.e., it is worse than using the standard \gls{ELBO} (and the gap with the \gls{ELBO} gets larger for increasing dimensionality $D$). Moreover, $\kernel_{\textrm{C-ISIR-DISIR}}$ alleviates the computational complexity of $\kernel_{\textrm{C-ISIR}}$, as measured by the number of \gls{MCMC} iterations it requires. \Cref{fig:bmnist_histograms_tau_D300} compares the histograms of the meeting time $\tau$ for both kernels; C-\acrshort{ISIR}-\acrshort{DISIR} requires significantly fewer iterations. (\Cref{fig:bmnist_histograms_tau_all} in \Cref{app:subsec:hist_meeting_time} shows that the histograms behave similarly across different values of $D$.)

\begin{table}[t]
    \centering
    \caption{Test log-likelihood for the \gls{VAE}. The unbiased estimators obtained via the coupled \acrshort{ISIR}-\acrshort{DISIR} kernel produce models with better predictive performance.%
    \label{tab:test_loglik}}
    \vspace*{-10pt}
    \subtable[Binarized MNIST.]{
        \setlength{\tabcolsep}{1.5pt}
        \label{tab:test_loglik_bmnist}
        \centering
        \scriptsize
        \begin{tabular}{cccc}
            \toprule
            & \multicolumn{3}{c}{dimensionality of $\lat$} \\
            & $20$ & $100$ & $300$ \\ \midrule
            \acrshort{ELBO} & $-90.05 \pm 0.21$ & $-89.96 \pm 0.14$ & $-90.63 \pm 0.12$ \\
            \acrshort{IWAE} & $-88.06 \pm 0.08$ & $-88.07 \pm 0.06$ & $-89.05 \pm 0.08$ \\ C-\acrshort{ISIR}-\acrshort{DISIR} & $\mathbf{-87.29 \pm 0.08}$ & $\mathbf{-86.75 \pm 0.10}$ & $\mathbf{-88.10 \pm 0.08}$ \\ \bottomrule
        \end{tabular}
    }
    \vspace*{-5pt}
    \subtable[Fashion-MNIST and CIFAR-10.]{
        \centering
        \scriptsize
        \label{tab:test_loglik_fmnist_cifar}
        \begin{tabular}{ccc}
            \toprule
            & Fashion-MNIST & CIFAR-10 \\ \midrule
            \acrshort{ELBO} & $-173.36 \pm 0.40$ & $-152.06 \pm 0.30$ \\
            \acrshort{IWAE} & $-170.50 \pm 0.30$ & $-149.72 \pm 0.39$ \\
            \acrshort{IWAE} + C-\acrshort{ISIR}-\acrshort{DISIR} & $\mathbf{-168.19 \pm 0.32}$& $\mathbf{-148.40 \pm 0.27}$ \\ \bottomrule
        \end{tabular}
    }
    \vspace*{-5pt}
\end{table}

The improved performance over \gls{IWAE} comes at the expense of computational complexity. The cost of \Cref{alg:couplingISIR} is roughly $10$ times the cost of computing $\hat{\nabla}_{\decp}\iwae$.

\parhead{Fashion-MNIST and CIFAR-10.}
The estimator based on \Cref{alg:couplingISIR} leads to improved models but it is also computationally more expensive. We now study the effect of switching to \Cref{alg:couplingISIR} after fitting a \gls{VAE} using the \gls{IWAE} objective. That is, we first fit the \gls{VAE} using the \gls{IWAE} objective for $200$ epochs, and then refine the result with the unbiased estimator based on $\kernel_{\textrm{C-ISIR-DISIR}}$.
We use two datasets, fashion-MNIST \citep{Xiao2017} and CIFAR-10 \citep{krizhevsky2009learning}, and set $D=100$.

\Cref{fig:fashionmnist_loglik} shows the train log-likelihood during optimization for fashion-MNIST. After switching from the \gls{IWAE} objective to the unbiased estimator of \Cref{alg:couplingISIR}, it improves. Additionally, \Cref{tab:test_loglik_fmnist_cifar} shows the test log-likelihood on both fashion-MNIST and CIFAR-10 after $300$ epochs. We can conclude that switching to an unbiased gradient estimator boosts the predictive performance of the \gls{VAE}.
\section{DISCUSSION}
\label{sec:discussion}

We have developed a practical algorithm to obtain unbiased estimators of the gradient of the log-likelihood for intractable models, and we have shown empirically that \glspl{VAE} fitted with unbiased estimators exhibit better predictive performance. Compared to \gls{ELBO} or \gls{IWAE} gradients, the main limitation of this approach is its higher computational cost and the fact that the running time is random. 
While one could obtain more accurate estimators simply by increasing the number of samples of the \gls{IWAE} bound, this would significantly increase the memory requirement, making it unpractical for datasets like CIFAR-10, and it would also remain biased for any (finite) number of samples.

The topic of coupling estimators is currently an active research field. We expect future work will improve the practical applicability of such estimators using methods like, e.g., control variates \citep{Craiu2020}.

\begin{acknowledgements} %
    We thank Andriy Mnih for his insightful comments and useful help. We also thank the anonymous reviewers and area chair for their feedback.
\end{acknowledgements}

\bibliography{references}

\clearpage

\appendix

\setcounter{equation}{15}
\setcounter{algocf}{4}
\setcounter{prop}{4}
\setcounter{figure}{4}

\section{DERIVATION OF THE COUPLING ESTIMATOR}
\label{app:sec:review_couplings}

Consider the expectation in \Cref{eq:expectation_h}. One choice to estimate the expectation is to build an \gls{MCMC} kernel $\kernel(\cdot\g u)$ that has $\pi(u)$ as its stationary distribution, run the Markov chain for some number of iterations $T$ to obtain samples $u^{(0)}, u^{(1)},\ldots, u^{(T)}$, and then approximate the expectation as $\E{\pi(u)}{h(u)}\approx \frac{1}{T-t_0+1}\sum_{t=t_0}^{T} h(u^{(t)})$, where the initial $t_0$ samples are thrown away as they are part of the burn-in period. However, this estimator is biased when the number of iterations $T$ is finite.\looseness=-1
Instead, \gls{MCMC} couplings provide an unbiased estimator in finite time \citep{glynn2014exact,jacob2017unbiased}. Coupling estimators use two \gls{MCMC} chains, each with invariant distribution $\pi(\cdot)$, which evolve according to a marginal transition kernel $\kernel(\cdot\g u)$ and a joint transition kernel $\kernel_{\textrm{C}}(\cdot,\cdot\g u,\coupl{u})$.
In our paper, we rely on a slight variation over the approach of \citet{jacob2017unbiased} proposed by \citet{VanettiDoucet2020}, which provides a construction also used by \citet{biswas2019estimating}.

Consider an integer $L\geq 1$. We draw the first Markov chain as $u^{(0)}\sim \pi_0(u)$ and $u^{(t)}\sim \kernel(u \g u^{(t-1)})$ for $t=1,\ldots,L$. We then draw $\coupl{u}^{(0)}$ (potentially conditionally upon $u^{(L-1)},u^{(L)}$) such that marginally $\coupl{u}^{(0)}\sim \pi_0(z)$. For $t> L$, we draw both states jointly as $u^{(t)}, \coupl{u}^{(t-L)} \sim \kernel_{\textrm{C}}(u, \coupl{u} \g u^{(t-1)}, \coupl{u}^{(t-L-1)})$. The meeting time is defined as $\tau=\inf\{t\geq L: u^{(t)}=\coupl{u}^{(t-L)}\}$.

We now provide an informal derivation of the estimator. First, we write the expectation of interest as
\begin{align}
    & \E{\pi(u)}{h(u)}=
    \lim_{N\rightarrow\infty}\frac{1}{L}\sum_{t=N-L+1}^N \mathbb{E}[h(u ^{(t)})]
    \nonumber\\
    & = \lim_{N\rightarrow\infty}\frac{1}{L}\Big\{\sum_{t=t_0}^{t_0+L-1}\mathbb{E}[h(u^{(t)})]+\sum_{t=t_0+L}^N \mathbb{E}[h(u^{(t)})] \nonumber\\
    &-\sum_{t=t_0}^{N-L} \mathbb{E}[h(u^{(t)})]\Big\}. 
\end{align}
Since $N\rightarrow\infty$, then the term within the limit can be equivalently rewritten as
\begin{align}
    & \frac{1}{L}\sum_{t=N-L+1}^N \mathbb{E}[h(u ^{(t)})]
    =
    \frac{1}{L}\Big\{\sum_{t=t_0}^{t_0+L-1}\mathbb{E}[h(u^{(t)})]\nonumber \\
    & +\sum_{t=t_0+L}^N \mathbb{E}[h(u^{(t)})]-\sum_{t=t_0+L}^{N} \mathbb{E}[h(u^{(t-L)})]\Big\}.
\end{align}
Taking into account that $u^{(t)}$ and $\coupl{u}^{(t)}$ have the same marginal distribution, then we can replace $u^{(t-L)}$ with $\coupl{u}^{(t-L)}$,
\begin{align}
    &\frac{1}{L}\sum_{t=N-L+1}^N \mathbb{E}[h(u ^{(t)})]
    =\frac{1}{L}\Big\{\sum_{t=t_0}^{t_0+L-1}\mathbb{E}[h(u^{(t)})] \nonumber\\
    &+\sum_{t=t_0+L}^N \mathbb{E}[h(u^{(t)})]
    -\sum_{t=t_0+L}^{N} \mathbb{E}[h(\coupl{u}^{(t-L)})]\Big\}.
\end{align}
We now combine the two sums in the right,
\begin{align}
    &\frac{1}{L}\sum_{t=N-L+1}^N \mathbb{E}[h(u ^{(t)})]
    =
    \frac{1}{L}\Big\{\sum_{t=t_0}^{t_0+L-1}\mathbb{E}[h(u^{(t)})]\nonumber\\
    &+\sum_{t=t_0+L}^N \mathbb{E}[h(u^{(t)})-h(\coupl{u}^{(t-L)})]\Big\}.
\end{align}
We now apply the fact that the two chains meet after some time $\tau$, i.e., $u^{(t)}=\coupl{u}^{(t-L)}$ for $t\geq \tau$. This gives
\begin{align}
    & \frac{1}{L}\sum_{t=N-L+1}^N \mathbb{E}[h(u ^{(t)})]
    =
    \frac{1}{L}\Big\{\sum_{t=t_0}^{t_0+L-1}\mathbb{E}[h(u^{(t)})] \nonumber\\
    & +\sum_{t=t_0+L}^{N\wedge(\tau-1)} \mathbb{E}[h(u^{(t)})-h(\coupl{u}^{(t-L)})]\Big\}.
\end{align}
When we consider the limit $N\rightarrow\infty$, the sum in the r.h.s.\ no longer depends on $N$; instead, it only contains (at most) $\tau-t_0-L$ terms,
\begin{align}
    &\E{\pi(u)}{h(u)}=
    \lim_{N\rightarrow\infty}\frac{1}{L}\sum_{t=N-L+1}^N \mathbb{E}[h(u ^{(t)})]
    \\
    &=
    \frac{1}{L}\Big\{\!\!\!\sum_{t=t_0}^{t_0+L-1}\!\!\!\!\mathbb{E}[h(u^{(t)})]+\!\!\sum_{t=t_0+L}^{\tau-1}\!\!\! \mathbb{E}[h(u^{(t)})-h(\coupl{u}^{(t-L)})]\Big\}.\nonumber
\end{align}
From this expression, we can obtain the unbiased estimator of $\E{\pi(u)}{h(u)}$ as given in \Cref{eq:estimator_vanetti}.

\section{MAXIMAL COUPLING KERNEL FOR CATEGORICAL DISTRIBUTIONS}
\label{app:sec:K_C_Cat}

Here we describe the maximal coupling kernel for two categorical distributions \citep{lindvall2002lectures}, which is used by the coupled \acrshort{DISIR} procedure from \Cref{subsec:coupledDISIR} (more specifically, in \Cref{alg_step:coupled_disir_sample_indicators} of \Cref{alg:C-DISIR}).

The kernel $\kernel_{\textrm{C-Cat}}$ is summarized in \Cref{alg:C-Cat}. It takes two (possibly unnormalized) probability vectors, $(w_1,\ldots,w_K)$ and $(v_1,\ldots,v_K)$, and returns a realization of two coupled categorical random variables $(\indic, \coupl{\indic})$.
\Cref{alg:C-Cat} is a maximal coupling scheme, in the sense that it achieves the theoretically maximum probability that $\indic=\coupl{\indic}$.

\section{PROOFS OF PROPOSITIONS AND LEMMAS}
\label{app:sec:proofs}

\subsection{Proposition \ref{prop:ISIR} and its Proof}
\label{app:subsec:proof_boundISIR}

The results in this section were derived by \citet[][Theorem 4]{Andrieu2010} and \citet[][Theorem 1]{andrieu2018uniform}, but we include them here for completeness.

\begin{prop}[\acrshort{ISIR} is invariant w.r.t.\ the posterior]
    \label{prop:ISIR}
    Under \Cref{ass:supportok}, for any $K\geq 2$, the \acrshort{ISIR} transition kernel $\kernel_{\text{\emph{ISIR}}}$ is invariant w.r.t.\ $p_{\decp,\encp}(\lat_{1:K}, \indic \g \obs)$ and the corresponding Markov chain is ergodic. Additionally, if \Cref{ass:weightsbounded} is satisfied, then for any initial value $(\lat_{1:K}^{(0)}, \indic^{(0)})$, the total variation distance w.r.t.\ the target is upper bounded by
    \begin{align}\label{eq:tv_bound_isir}
        \left|\left| \kernel_{\text{\emph{ISIR}}}^{T}(\cdot,\cdot \g \lat_{1:K}^{(0)}, \indic^{(0)}) - p_{\decp,\encp}(\cdot,\cdot \g \obs) \right|\right|_{\textrm{TV}}
        &\leq \rho_K^T,
    \end{align}
    where $\rho_K:=1 - \frac{K-1}{2 w_{\decp,\encp}^{\textrm{max}}/p_{\decp}(\obs) + K-2}<1$,
    and the notation $\kernel_{\text{\emph{ISIR}}}^{T}(\cdot,\cdot \g \lat_{1:K}^{(0)}, \indic^{(0)})$ indicates the distribution of the state of the chain after $T$ steps of the kernel initialized at $(\lat_{1:K}^{(0)}, \indic^{(0)})$.
\end{prop}

We now prove \Cref{prop:ISIR}.
Under \Cref{ass:supportok}, the extended target distribution
\begin{equation}
    \label{eq:p_augmented}
    p_{\decp,\encp}(\lat_{1:K},\indic \g \obs) = 
    \frac{1}{K} \; p_{\decp}(\lat_{\indic} \g \obs) \prod_{k=1,k\neq \indic}^{K} q_{\encp}(\lat_k \g \obs)
\end{equation}
is well-defined.
The transition kernel of \acrshort{ISIR} is defined by \Cref{alg:ISIR} and given by
\begin{align}\label{eq:KernelISIRjoint}
    &\kernel_{\textrm{ISIR}}(\lat_{1:K}^\star, \indic^\star \g \lat_{1:K}, \indic)=\sum_{\indic_{\textrm{aux}}=1}^K\frac{1}{K}\delta_{\lat_\indic}(\lat^\star_{\indic_{\textrm{aux}}}) \nonumber \\
    & \times \left(\prod_{k=1,k\neq\indic_{\textrm{aux}}}^K q(\lat^\star_k \g \obs)\right) \frac{w_{\decp,\encp}(\lat^{\star}_{\indic^\star})}{\sum_{k=1}^K w_{\decp,\encp}(\lat^{\star}_{k})}.
\end{align}

To prove invariance, the marginalization $\sum_{\indic=1}^K \int p_{\decp,\encp}(\lat_{1:K},\indic \g \obs) \kernel_{\textrm{ISIR}}(\lat_{1:K}^\star, \indic^\star \g \lat_{1:K}, \indic)\textrm{d}\lat_{1:K}$ should be equal to $p_{\decp,\encp}(\lat^{\star}_{1:K},\indic^{\star} \g \obs)$. Indeed,
\begin{align}
    &\sum_{\indic=1}^K \int p_{\decp,\encp}(\lat_{1:K},\indic \g \obs) \kernel_{\textrm{ISIR}}(\lat_{1:K}^\star, \indic^\star \g \lat_{1:K}, \indic)\textrm{d}\lat_{1:K} \nonumber\\
    &=\sum_{\indic_{\textrm{aux}}=1}^K  p_{\decp,\encp}(\lat^{\star}_{1:K},\indic_{\textrm{aux}} \g \obs) p_{\decp,\encp}(\indic^{\star} \g \lat^{\star}_{1:K}, \obs)\nonumber\\
    &= p_{\decp,\encp}(\lat^{\star}_{1:K},\indic^{\star} \g \obs).
\end{align}
Here, we have first integrated out the latent variables $\lat_{1:K}$ except the $\indic$-th one. Then, we have integrated out $\lat_{\indic}$ applying the properties of the Dirac delta function; this makes the resulting expression independent of $\indic$ and therefore we can easily get rid of the sum over $\indic$. Next, we have recognized the term $p_{\decp,\encp}(\lat_{1:K}^\star,\indic_{\textrm{aux}} \g \obs)$ (from \Cref{eq:p_augmented}), and we have applied that $p_{\decp,\encp}(\indic^{\star} \g \lat^{\star}_{1:K}, \obs)$ is a categorical distribution with probability proportional to $w_{\decp,\encp}(\lat^{\star}_{\indic^{\star}})$. Finally, we have marginalized out $\indic_{\textrm{aux}}$, leading to the final expression.

\begin{algorithm}[t]
    \SetAlgoLined
    \DontPrintSemicolon
    \caption{%
        Maximal coupling kernel for categoricals, $\kernel_{\textrm{C-Cat}}(\cdot, \cdot \g (w_1,\ldots,w_K), (v_1,\ldots,v_K) )$ %
        \label{alg:C-Cat}
    }
    \SetKwComment{mycomment}{$\triangleright$ }{}
    \KwIn{Two unnormalized probability vectors $(w_1,\ldots,w_K)$ and $(v_1,\ldots,v_K)$}
    \KwOut{A sample $\indic, \coupl{\indic}$ from the maximal coupling kernel}
    Normalize the input vectors, obtaining $\widetilde{w}_k\propto w_k$ and $\widetilde{v}_k \propto v_k$ for $k=1,\ldots, K$\;
    Compute the total variation $\gamma = \frac{1}{2}\sum_{k=1}^{K}|\widetilde{w}_k - \widetilde{v}_k|$ \;
    Sample $u\sim \textrm{Uniform}(0, 1)$\;
    \uIf(coupling occurs){$u\leq 1-\gamma$}{
        Sample $\indic \sim \textrm{Cat}(p_1, \ldots, p_K)$ with $p_k\propto \min(\widetilde{w}_k,\widetilde{v}_k)$\;
        Return $(\indic, \indic)$\;
    }\Else(coupling does not occur){
        Sample $\indic \sim \textrm{Cat}(p_1, \ldots, p_K)$ with $p_k\propto \max(\widetilde{w}_k-\widetilde{v}_k,0)$\;
        Sample $\coupl{\indic} \sim \textrm{Cat}(p_1, \ldots, p_K)$ with $p_k\propto\max(\widetilde{v}_k-\widetilde{w}_k,0)$\;
        Return $(\indic, \coupl{\indic})$\;
    }
\end{algorithm}

This transition kernel is $\phi$-irreducible and aperiodic under \Cref{ass:supportok}; therefore, the Markov chain is ergodic \citep{tierney1994markov}.

When simulating a Markov chain $(z^{(t)}_{1:K},\indic^{(t)})_{t\geq0}$ according to $\kernel_{\textrm{ISIR}}$, the $\indic^{(t)}$-th latent variable, i.e., $(z^{(t)}:=z^{(t)}_{\indic^{(t)}})_{t\geq0}$, is also a Markov chain with the transition kernel originally described by \citet{Andrieu2010}, which we denote $\kernel_{\textrm{ISIR,orig}}$. We can obtain this kernel from \Cref{eq:KernelISIRjoint} by setting $\lat_{\indic} = \lat$, $\lat^\star_{\indic^\star} = \lat^\star$, and marginalizing out the variables $\indic^\star$ and $\lat_{1:K}^\star$. This gives
\begin{align}\label{eq:KernelISIRmarginal}
    &\kernel_{\textrm{ISIR,orig}}(\lat^{\star} \g \lat)
     =
    \frac{1}{K} \sum_{\indic_{\textrm{aux}}=1}^{K} \sum_{\indic^\star=1} ^{K} \int \delta_{\lat}( \lat_{\indic_{\textrm{aux}}}^{\star}) \\
    & \;\;\;\times \!\!\left(\! \prod_{k\neq \indic_{\textrm{aux}}} \!\! q_{\encp}(\lat^{\star}_k \g \obs)\!\! \right)\! \frac{w_{\decp,\encp}(\lat^{\star}_{\indic^\star})}{\sum_{k=1}^K w_{\decp,\encp}(\lat^{\star}_{k})} \delta_{\lat^{\star}_{\indic^\star}}(\lat^{\star}) \textrm{d}\lat^{\star}_{1:K} \nonumber\\
    & = 
    \sum_{\indic^\star=1}^{K} \int \delta_{\lat}( \lat^{\star}_1) \left( \prod_{k=2}^{K} q_{\encp}(\lat^{\star}_k \g \obs) \right) \nonumber \\ 
    & \;\;\;\times \frac{w_{\decp,\encp}(\lat^{\star}_{\indic^\star})}{\sum_{k=1}^K w_{\decp,\encp}(\lat^{\star}_{k})} \delta_{\lat^{\star}_{\indic^\star}}(\lat^{\star}) \textrm{d}\lat^{\star}_{1:K},\nonumber
\end{align}
where we have used the symmetry of the kernel w.r.t.\ $\indic_{\textrm{aux}}$ and have arbitrarily considered the term with $\indic_{\textrm{aux}}=1$.

We next prove the bound on the total variation distance.
Given that each term is non-negative, we can lower bound $\kernel_{\textrm{ISIR,orig}}(\lat^{\star} \g \lat)$ by getting rid of the term corresponding to $\indic^\star=1$ from the summation. This gives
\begin{align}
    &\kernel_{\textrm{ISIR,orig}}(\lat^{\star} \g \lat) \geq \sum_{\indic^\star=2}^{K} \int \delta_{\lat}( \lat^{\star}_1) \left( \prod_{k=2}^{K} q_{\encp}(\lat^{\star}_k \g \obs) \right)  \\
    & \times \frac{w_{\decp,\encp}(\lat^{\star}_{\indic^\star})}{\sum_{k=1}^K  w_{\decp,\encp}(\lat^{\star}_{k})} \delta_{\lat^{\star}_{\indic^\star}}(\lat^{\star}) \textrm{d}\lat^{\star}_{1:K}.\nonumber
\end{align}
Using the definition of the importance weights, $w_{\decp,\encp}(\lat) = p_{\decp}(\obs, \lat) / q_{\encp}(\lat \g \obs) = p_{\decp}(\obs) p_{\decp}(\lat \g \obs) / q_{\encp}(\lat \g \obs)$, this yields
\begin{align}
    &\kernel_{\textrm{ISIR,orig}}(\lat^{\star} \g \lat) \geq \sum_{\indic^\star=2}^{K} \int \delta_{\lat}( \lat^{\star}_1) \left( \prod_{k=2,k\neq \indic^{\star}}^{K} q_{\encp}(\lat^{\star}_k \g \obs) \right)\nonumber\\
    & \times \frac{p_{\decp}(\obs) p_{\decp}(\lat^{\star}_{\indic^\star}\g \obs)}{\sum_{k=1}^K w_{\decp,\encp}(\lat^{\star}_{k})} \delta_{\lat^{\star}_{\indic^\star}}(\lat^{\star}) \textrm{d}\lat^{\star}_{1:K}.
\end{align}
By \Cref{ass:weightsbounded}, we have $w_{\decp,\encp}(\lat^{\star}_{1})+w_{\decp,\encp}(\lat^{\star}_{2})\leq 2 w_{\decp,\encp}^{\textrm{max}}$, and we can further lower bound $\kernel_{\textrm{ISIR,orig}}$ as
\begin{align}
    &\kernel_{\textrm{ISIR,orig}}(\lat^{\star} \g \lat) \geq \sum_{\indic^\star=2}^{K} \int \delta_{\lat}( \lat^{\star}_1) \left( \prod_{k=2,k\neq \indic^{\star}}^{K} q_{\encp}(\lat^{\star}_k \g \obs) \right)\nonumber\\
    &\times \frac{p_{\decp}(\obs) p_{\decp}(\lat^{\star}_{\indic^\star}\g \obs)}{2w_{\decp,\encp}^{\textrm{max}} + \sum_{k=3}^K w_{\decp,\encp}(\lat^{\star}_{k})} \delta_{\lat^{\star}_{\indic^\star}}(\lat^{\star}) \textrm{d}\lat^{\star}_{1:K}.
\end{align}
Next, by using the symmetry of the integrand w.r.t.\ $\indic^{\star}$, it follows that
\begin{equation}
    \kernel_{\textrm{ISIR,orig}}(\lat^{\star} \g \lat)\geq \E{}{\frac{(K-1) p_{\decp}(\obs) p_{\decp}(\lat^{\star} \g \obs)} {2w_{\decp,\encp}^{\textrm{max}} + \sum_{k=3}^{K} w_{\decp,\encp}(\lat^{\star}_k)}},
\end{equation}
where the expectation is w.r.t.\ $z^{\star}_k \sim q_{\encp}(\cdot \g \obs)$ for $k=3,...,K$. We finally apply Jensen's inequality, $\E{q}{f(\cdot)} \geq f(\E{q}{\cdot}$) for the convex function $f(x)=1/x$, obtaining
\begin{align}
   & \kernel_{\textrm{ISIR,orig}}(\lat^{\star} \g \lat) \geq \frac{(K-1) p_{\decp}(\obs) p_{\decp}(\lat^{\star} \g \obs)} {\E{}{2w_{\decp,\encp}^{\textrm{max}} + \sum_{k=3}^{K} w_{\decp,\encp}(\lat^{\star}_k)}} \nonumber \\
   & = \frac{(K-1)p_{\decp}(\obs)}{2w_{\decp,\encp}^{\textrm{max}} + (K-2) p_{\decp}(\obs) } p_{\decp}(\lat^{\star} \g \obs).
\end{align}
This kernel thus satisfies a minorization condition and thus
\begin{equation}\label{eq:tv_bound_isir2}
     \left|\left| \kernel_{\textrm{ISIR,orig}}^{T}(\cdot \g \lat^{(0)}) - p_{\decp}(\cdot \g \obs) \right|\right|_{\textrm{TV}}
 \leq \rho_K^T,
\end{equation}
where $\rho_K:=1 - \frac{K-1}{2 w_{\decp,\encp}^{\textrm{max}}/p_{\decp}(\obs) + K-2}$.

The bound in \Cref{eq:tv_bound_isir} follows directly, since $z^{(t)}:=z^{(t)}_{\indic^{(t)}}$ and, under the transition kernel $\kernel_{\textrm{ISIR}}$, the remaining variables are sampled from the full conditional distribution of the extended target from \Cref{eq:p_augmented}, so we have for any $T\geq 0$,
\begin{align}
    &\left|\left| \kernel_{\textrm{ISIR,orig}}^{T}(\cdot \g \lat^{(0)}) - p_{\decp}(\cdot \g \obs) \right|\right|_{\textrm{TV}} \\
    & =\left|\left| \kernel_{\textrm{ISIR}}^{T}(\cdot, \cdot \g \lat_{1:K}^{(0)}, \indic^{(0)}) - p_{\decp,\encp}(\cdot, \cdot \g \obs) \right|\right|_{\textrm{TV}}.\nonumber
\end{align}

\subsection{Proof of Proposition \ref{prop:DISIR}}
\label{app:subsec:proof_invarianceDISIR}

We now prove here that the \acrshort{DISIR} kernel admits $p^{\textrm{DISIR}}_{\decp,\encp}(\xi_{1:K}, \indic \g \obs)$ defined in \Cref{eq:augmentedtargetcorrelated} as invariant distribution. The transition kernel $\kernel_{\textrm{DISIR}}(\cdot, \cdot \g \xi_{1:K}, \indic)$ is defined through \Cref{alg:DISIR} and can be written as
\begin{align}
    &\kernel_{\textrm{DISIR}}(\xi_{1:K}^\star, \indic^\star \g \xi_{1:K}, \indic)=\sum_{\indic_{\textrm{aux}}=1}^K\frac{1}{K}\delta_{\xi_\indic}(\xi^\star_{\indic_{\textrm{aux}}})\\
    & \times\! \prod_{k=1}^{\indic_{\textrm{aux}}-1} \!\!p_\corr(\xi^\star_{k} \g \xi^\star_{k+1}) \!\!\!\prod_{k=\indic_{\textrm{aux}}+1}^K \!\!\!p_\corr(\xi^\star_{k} \g \xi^\star_{k-1})
    \frac{w_{\decp,\encp}(\lat^{\star}_{\indic^\star})}{\sum_{k=1}^K w_{\decp,\encp}(\lat^{\star}_{k})},\nonumber
\end{align}
for $z^{\star}_k=g_{\encp}(\xi^{\star}_k,\obs)$.

For the kernel to be invariant, the marginalization
$\sum_{\indic=1}^K \int p^{\textrm{DISIR}}_{\decp,\encp}(\xi_{1:K},\indic \g \obs) \kernel_{\textrm{DISIR}}(\xi_{1:K}^\star, \indic^\star \g \xi_{1:K}, \indic)\textrm{d}\xi_{1:K}$
should be equal to $p^{\textrm{DISIR}}_{\decp,\encp}(\xi_{1:K}^\star,\indic^\star \g \obs)$. We obtain this marginalization below. We first define
\begin{equation}
    \label{eq:definition_nu}
    p^{\textrm{DISIR}}_{\decp,\encp}(\xi \g \obs):=\frac{w_{\decp,\encp}( g_{\encp}(\xi,\obs))q(\xi)}{p_{\decp}(x)}.
\end{equation}
(\Cref{eq:definition_nu} gives the marginal distribution of $\xi_{\indic}$ obtained after integrating out the rest of latent variables from \Cref{eq:augmentedtargetcorrelated}.)
Similarly to the proof in \Cref{app:subsec:proof_boundISIR}, we first integrate out the variables $\xi_{1:K}$ except the $\indic$-th one, and then we marginalize out $\xi_\indic$ taking into account the integration property of the Dirac delta function; this allows us to get rid of the sum over $\indic$. Specifically, we have
\begin{align}
    &\sum_{\indic=1}^K \int p^{\textrm{DISIR}}_{\decp,\encp}(\xi_{1:K},\indic \g \obs) \kernel_{\textrm{DISIR}}(\xi_{1:K}^\star, \indic^\star \g \xi_{1:K}, \indic)\textrm{d}\xi_{1:K} \nonumber\\
    &=\sum_{\indic=1}^K \sum_{\indic_{\textrm{aux}}=1}^K\frac{1}{K} \int \frac{p_{\decp,\encp}^{\textrm{DISIR}}(\xi_{\indic} \g \obs)}{K} \delta_{\xi_\indic}(\xi^\star_{\indic_{\textrm{aux}}}) \nonumber \\
    & \;\;\;\times\!\!\!
    \prod_{k=1}^{\indic_{\textrm{aux}}-1} \!\!p_\corr(\xi^\star_{k} \g \xi^\star_{k+1}) \!\!\!\!\prod_{k=\indic_{\textrm{aux}}+1}^K \!\!\!\!\! p_\corr(\xi^\star_{k} \g \xi^\star_{k-1})
    \frac{w_{\decp,\encp}(\lat^{\star}_{\indic^\star})}{\sum_{k=1}^K w_{\decp,\encp}(\lat^{\star}_{k})} \textrm{d}\xi_{\indic}\nonumber\\
    &=\sum_{\indic_{\textrm{aux}}=1}^K  p^{\textrm{DISIR}}_{\decp,\encp}(\xi^{\star}_{1:K},\indic_{\textrm{aux}} \g \obs) \frac{w_{\decp,\encp}(\lat^{\star}_{\indic^\star})}{\sum_{k=1}^K w_{\decp,\encp}(\lat^{\star}_{k})}\nonumber\\
    &=\sum_{\indic_{\textrm{aux}}=1}^K  p^{\textrm{DISIR}}_{\decp,\encp}(\xi^{\star}_{1:K},\indic_{\textrm{aux}} \g \obs)p_{\decp,\encp}(\indic^{\star} \g \obs, \xi_{1:K})\nonumber\\
    &= p^{\textrm{DISIR}}_{\decp,\encp}(\xi^{\star}_{1:K},\indic^\star \g \obs).
\end{align}
Here, we have additionally recognized the term $p^{\textrm{DISIR}}_{\decp,\encp}(\xi^{\star}_{1:K},\indic_{\textrm{aux}} \g \obs)$ (see \Cref{eq:augmentedtargetcorrelated}). For the second-to-last step, we have applied the fact that the conditional distribution of $\indic$ under \Cref{eq:augmentedtargetcorrelated}, $p^{\textrm{DISIR}}_{\decp,\encp}(\indic \g \obs, \xi_{1:K})$, is a categorical with probability proportional to $w_{\decp,\encp}(g_{\encp}(\xi_\indic,x))$ (this posterior is analogous to the \acrshort{ISIR} case). To establish this result, we note that
\begin{align}
    & p^{\textrm{DISIR}}_{\decp,\encp}(\indic \g \obs, \xi_{1:K}) \propto w_{\decp,\encp}(g_{\encp}(\xi_\indic,x))q(\xi_\indic) \\
    & \times \prod_{k=1}^{\indic-1} p_\corr(\xi_{k} \g \xi_{k+1}) \prod_{k=\indic+1}^K p_\corr(\xi_{k} \g \xi_{k-1}).\nonumber
\end{align}
Since $p_\corr$ is reversible with respect to $q$, it follows that
\begin{align}
    &q(\xi_\indic) \prod_{k=1}^{\indic-1} p_\corr(\xi_{k} \g \xi_{k+1}) \prod_{k=\indic+1}^K p_\corr(\xi_{k} \g \xi_{k-1}) \nonumber \\
    & = q(\xi_1)\prod_{k=2}^{K}  p_\corr(\xi_{k} \g \xi_{k-1}),
\end{align}
so this product is independent of $\indic$ and we have
\begin{equation}
    \label{eq:conditionalindicdependent}
    p^{\textrm{DISIR}}_{\decp,\encp}(\indic \g \obs, \xi_{1:K})\propto w_{\decp,\encp}(g_{\encp}(\xi_\indic,x)).
\end{equation}
This establishes the proof of invariance.
The \acrshort{DISIR} kernel is ergodic for the same reasons as \acrshort{ISIR}.

\subsection{Proof of Proposition \ref{lemma:invariance}}
\label{app:subsec:prooflemmaaugmentatedtarget}

We prove here \Cref{lemma:invariance}. From the \acrshort{DISIR} invariant distribution in \Cref{eq:augmentedtargetcorrelated}, it follows that the marginal distribution of $\xi_{\indic}$ is given by $p_{\decp,\encp}^{\textrm{DISIR}}(\xi_\indic\g\obs)$ (\Cref{eq:definition_nu}).

We need to show that for $\xi \sim p_{\decp,\encp}^{\textrm{DISIR}}(\xi \g \obs)$, then $z:=g_{\encp}(\xi,x) \sim p_{\decp}(\lat \g \obs)$. For any test function $f(\cdot)$,
\begin{align}
    &\E{p^{\textrm{DISIR}}_{\decp,\encp}(\xi \g \obs)}{f(g_{\encp}(\xi,\obs))} \\
    & =\int f(g_{\encp}(\xi,\obs)) \frac{w_{\decp,\encp}( g_{\encp}(\xi,\obs))q(\xi)}{p_{\decp}(\obs)}\textrm{d}\xi \nonumber\\
    & =\int f(g_{\encp}(\xi,\obs)) \frac{p_{\decp}(\obs, g_{\encp}(\xi,\obs))}{q_{\encp}(g_{\encp}(\xi,\obs) \g \obs)}\frac{q(\xi)}{p_{\decp}(\obs)}\textrm{d}\xi.\nonumber
\end{align}
Under \Cref{ass:reparamtrick}, we know that if $\xi \sim q(\xi)$ then $\lat = g_{\encp}(\xi,\obs)\sim q_{\encp}(\lat \g \obs)$. Thus, by using the change of variables $\lat=g_{\encp}(\xi,\obs)$, we have
\begin{align}
    & \E{p^{\textrm{DISIR}}_{\decp,\encp}(\xi \g \obs)}{f(g_{\encp}(\xi,\obs))} =\int f(\lat) \frac{p_{\decp}(\obs, \lat)}{q_{\encp}(\lat \g \obs)}\frac{q_{\encp}(\lat \g \obs)}{p_{\decp}(\obs)}\textrm{d}\lat \nonumber\\ 
    & =\E{p_{\decp}(\lat \g \obs)}{f(\lat)}.
\end{align}
This completes the proof of the first part of \Cref{lemma:invariance}.

The second part says that, for $\corr=0$, the variables $(\lat_{1:K}, \indic)$ are distributed according to the augmented posterior. In this case, the variables $\xi_k$ for $k \neq \indic$ are independent and identically distributed according to $q(\xi)$. Thus, if $(\xi_{1:K},\indic) \sim p^{\textrm{DISIR}}_{\decp,\encp}(\xi_{1:K},\indic \g \obs)$ then, for $z_k=g_{\encp}(\xi_k,\indic)$, we have $(z_{1:K},\indic) \sim p_{\decp,\encp}(\lat_{1:K},\indic \g \obs)$.

\subsection{Proof of Proposition \ref{prop:identityexpectation}}
\label{app:subsec:proof_RaoBlackwellestimators}

We establish here the identity in Eq~10. This is a generalization of Theorem 6 in \citet{Andrieu2010}. From the result in \Cref{lemma:invariance} and the definition of $p^{\textrm{DISIR}}_{\decp,\encp}(\xi_{1:K}, \indic \g \obs)$ (\Cref{eq:augmentedtargetcorrelated}), it follows that
\begin{align}
    & \int h(\lat) p_{\decp}(\lat \g \obs)\textrm{d}\lat
    \\
    & = \sum_{\indic=1}^K \int h(g_\phi(\xi_\indic,\obs)) p^{\textrm{DISIR}}_{\decp,\encp}(\xi_{1:K}, \indic \g \obs) \textrm{d}\xi_{1:K}\nonumber\\
    &=\!\!\int \!\! \Big[\!\sum_{\indic=1}^K h(g_\phi(\xi_{\indic},\obs)) p^{\textrm{DISIR}}_{\decp,\encp}(\indic \g \obs, \xi_{1:K}) \!\Big] p^{\textrm{DISIR}}_{\decp,\encp}(\xi_{1:K} \g \obs) \textrm{d}\xi_{1:K}\nonumber\\
    &=\!\!\int \Big[\sum_{\indic=1}^K \widetilde{w}_{\decp,\encp}^{(\indic)} h(g_\phi(\xi_{\indic},\obs))  \Big] p^{\textrm{DISIR}}_{\decp,\encp}(\xi_{1:K} \g \obs) \textrm{d}\xi_{1:K},\nonumber
\end{align}
where $\widetilde{w}_{\decp,\encp}^{(\indic)} \propto w_{\decp,\encp}(g_{\encp}( \xi_{\indic}, \obs))$ are the normalized importance weights. The last equality follows from \Cref{eq:conditionalindicdependent} used in the proof of \Cref{app:subsec:proof_invarianceDISIR}. The result thus follows.

\subsection{Proof of Proposition \ref{prop:unbiasedcond}}
\label{app:subsec:proof_unbiased_cond}

The following shows that the conditions established by \citet{middleton2018unbiasedEJS} to establish the fact that the estimator of \cite{jacob2017unbiased} can be computed in finite expected time and admit a finite variance are also applicable to the estimator of \Cref{eq:estimator_vanetti}. The proof follows the approach of \citet[][Proposition 3.1]{jacob2017unbiased} and \citet[][Theorem 1]{middleton2018unbiasedEJS} but some details differ.

Here, we use the notation $\mu(h):=\int h(u) \mu(u)\textrm{d}u$ for any test function $h(u)$ and probability density $\mu(u)$. Our goal is to estimate $H:=\pi(h)$.
Firstly, by condition (\ref{item:finite_tau}), we have $\E{}{\tau}<\infty$, so the estimator $\hat{H}:=\hat{\pi}(h)$ from \Cref{eq:estimator_vanetti} can be computed in finite expected time.

Now let us denote by $L_{2}$ the complete space of random variables with finite second moment. We consider the sequence of random variables $(\hat{\pi}_N(h))_{N\geq k+L}$ defined by 
\begin{align}
    \hat{\pi}_N(h) &  =\frac{1}{L}\left(\sum_{t=k}^{k+L-1} h(u^{(t)})
    + \!\!\!\sum_{t=k+L}^{N} h(u^{(t)})-h(\coupl{u}^{(t-L)})\!\!\right) \nonumber \\
    & =\frac{1}{L} \sum_{t=k}^{N} \Delta_{t},
\end{align}
where $\Delta_{t}:=h(u^{(t)})-h(\coupl{u}^{(t-L)})$ for $t\geq k+L$ and $\Delta_{t}:=h(u^{(t)})$ for $k\leq t<k+L$. 
We next show that this sequence is a Cauchy sequence in $L_{2}$ converging to $\hat{\pi}(h)$.

As $\E{}{\tau}<\infty$, we have $\mathbb{P}(\tau<\infty)=1$ and $u^{(t)}=\coupl{u}^{(t-L)}$ for $t\geq\tau$ under condition (\ref{item:chains_stay_together}). Thus, it follows that $\hat{\pi}_N(h) \rightarrow \hat{\pi}(h)$
almost surely. For positive integers $N, N'$ such that $k+L\leq N< N'$, we have
\begin{align}
     & \E{}{\left(\hat{\pi}_N(h)-\hat{\pi}_{N'}(h)\right)^{2}} 
     =\frac{1}{L^2}\sum_{s=N+1}^{N'}\sum_{t=N+1}^{N'}\E{}{\Delta_{s}\Delta_{t}}\nonumber \\
     & \le \frac{1}{L^2}  \sum_{s=N+1}^{N'} \sum_{t=N+1}^{N'}  \E{}{\Delta_{s}^{2}}^{1/2}  \E{}{\Delta_{t}^{2}}^{1/2} \nonumber \\
     & =\frac{1}{L^2}\left(\sum_{t=N+1}^{N'}\E{}{\Delta_{t}^{2}}^{1/2}\right)^{2}.
\end{align}
Since $\E{}{\Delta_{t}^{2}}=\E{}{\Delta_{t}^{2}\mathbb{I}_{\tau>t}}$, where $\mathbb{I}$ is the indicator function, by Holder's inequality we have
\begin{align}
    \E{}{\Delta_{t}^{2}} &\le\E{}{\left|\Delta_{t}\right|^{2+\eta}}^{\frac{1}{1+\frac{\eta}{2}}}\E{}{\mathbb{I}_{\tau>t}}^{\frac{\eta}{2+\eta}} \nonumber \\
    & \le D{}^{\frac{1}{1+\frac{\eta}{2}}}\mathbb{P}(\tau>t){}^{\frac{\eta}{2+\eta}},
\end{align}
where $\E{}{\left|\Delta_{t}\right|^{2+\eta}}<D$ for all $t$
as $\E{}{|h(u^{(t)})|^{2+\eta}}<D$ by condition (\ref{item:finite_high_order_moment}). Consequently, we have
\begin{align}
  & \E{}{\left(\hat{\pi}_N(h)-\hat{\pi}_{N'}(h)\right)^{2}} \\
  & \le \frac{1}{L^2}\left(\sum_{t=N+1}^{N'}\left(D^{\frac{1}{1+\frac{\eta}{2}}}\mathbb{P}(\tau>t){}^{\frac{\eta}{2+\eta}}\right)^{\frac{1}{2}}\right)^{2} \nonumber \\
  & = \frac{1}{L^2} D^{\frac{1}{1+\frac{\eta}{2}}}\left(\sum_{t=N+1}^{N'}\mathbb{P}(\tau>t){}^{\frac{1}{2}\frac{\eta}{2+\eta}}\right)^{2}.\nonumber 
\end{align}
Defining $\lambda:=\frac{1}{2}\frac{\eta}{2+\eta}$, it follows from condition (\ref{item:finite_tau}) that $\mathbb{P}(\tau>t)\le Ct^{-\kappa}$ for $\kappa>1/\lambda$,
which yields
\begin{align}
    \sum_{t=N+1}^{\infty}\mathbb{P}(\tau>t)^{\lambda} & \le C\sum_{t=N+1}^{\infty}\frac{1}{t^{\lambda\kappa}}<\infty.
\end{align}
Thus, we have $\lim_{N\rightarrow\infty}\sum_{t=N+1}^{\infty}\mathbb{P}(\tau>t)^{\lambda}=0$.
Hence, we have proved $\hat{\pi}_N(h)$ is a Cauchy sequence in $L_{2}$, and has finite first and second moments, so $\hat{\pi}(h)$ has finite variance. As Cauchy sequences are bounded, the dominated convergence theorem shows that
\begin{equation}
    \E{}{\hat{\pi}(h)}=\E{}{\lim_{N\rightarrow \infty} \hat{\pi}_N(h)}=\lim_{N\rightarrow \infty} \E{}{\hat{\pi}_N(h)},
\end{equation}
and, under condition (\ref{item:convergence_markov_chain}), we have
\begin{equation}
    \lim_{N\rightarrow \infty} \E{}{\hat{\pi}_N(h)}= \lim_{N\rightarrow \infty} \frac{1}{L}\sum_{t=N-L+1}^N \E{}{h(u^{(t)})}=\pi(h).
\end{equation}

\subsection{Proof of Proposition \ref{prop:estimateok}}
\label{app:subsec:proof_properties_unbiased_estimator_disir}

To prove \Cref{prop:estimateok}, we need to check that the conditions (\ref{item:convergence_markov_chain}) to (\ref{item:chains_stay_together}) of \Cref{prop:unbiasedcond} are satisfied for the coupled \acrshort{ISIR}-\acrshort{DISIR} kernel from \Cref{alg:C-DISIR}. Condition (\ref{item:convergence_markov_chain}) is satisfied as the \acrshort{ISIR} kernel is $\phi$-irreducible and aperiodic \citep[see, e.g.,][]{tierney1994markov}. Condition (\ref{item:finite_high_order_moment}) is satisfied by assumption. Condition (\ref{item:chains_stay_together}) is also satisfied by design of \Cref{alg:C-DISIR}---once the chains are coupled, they remain equal to each other forever. We next check that condition (\ref{item:finite_tau}) is also satisfied.

We recall that the transition kernel we couple is a composition of the \acrshort{ISIR} kernel followed by the \acrshort{DISIR} kernel. Here we show that, at each iteration, the coupled \acrshort{ISIR} kernel couples with a probability that is lower bounded by a quantity strictly positive independent of the current states of the two chains. This ensures that the distribution of the meeting time $\tau$ has tails decreasing geometrically fast.

As discussed in \Cref{subsec:coupledDISIR}, the coupled \acrshort{ISIR} kernel from \Cref{alg:C-DISIR} couples when (i) both indicators sampled in \Cref{alg_step:coupled_disir_sample_indicators} are equal, i.e., $\indic^\star=\coupl{\indic}^\star$, and (ii) these indicators are different from $\indic_{\textrm{aux}}$. Event (i) is driven by the joint kernel $\kernel_{\textrm{C-Cat}}$, whose probability of coupling is $1-\gamma$, where $\gamma$ is defined in \Cref{alg:C-Cat}. The probability of event (ii) is equal to the probability that the sampled indicators are different from  $\indic_{\textrm{aux}}$, which we assume equal to 1 (without loss of generality).

When we use \Cref{alg:C-Cat} to couple the \acrshort{ISIR} chains, we have $w_k = w_{\decp,\encp}(\lat^{\star}_k)$ and $v_k = w_{\decp,\encp}(\bar{\lat}^{\star}_k)$ (see \Cref{alg_step:coupled_disir_sample_indicators} of \Cref{alg:C-DISIR}). Thus, the unnormalized weights $(w_k)_{k=1,\ldots,K}$ and $(v_k)_{k=1,\ldots,K}$ before coupling differ at most by a single entry, which we assumed above to be the first entry, i.e., $w_1 \neq v_1$ and $w_k = v_k$ for $k=2,\ldots,K$. The normalized probabilities are thus
\begin{equation}
    \widetilde{w}_k = \frac{w_k}{w_1 + S},\quad \widetilde{v}_k=\frac{v_k}{v_1 + S},
\end{equation}
where
\begin{equation}
    S=\sum_{k=2}^{K} v_k=\sum_{k=2}^{K} w_k.
\end{equation}

Therefore, the probability of coupling of \acrshort{ISIR} is
\begin{align}
    P_{\textrm{meet}} \geq \E{}{(1-\gamma) \frac{\sum_{k=2}^K \min(\widetilde{w}_k, \widetilde{v}_k)}{\sum_{k=1}^K \min(\widetilde{w}_k,\widetilde{v}_k)}},
\end{align}
where the expectation is w.r.t.\ to the joint distribution of the two chains at time $t$.
Using the identity $|a-b|=a+b-2\min(a,b)$, the term $1-\gamma$ can be simplified as
\begin{align}
    1-\gamma =
    1 - \frac{1}{2} \sum_{k=1}^K |\widetilde{w}_k - \widetilde{v}_k|
    = \sum_{k=1}^K \min(\widetilde{w}_k,\widetilde{v}_k).
\end{align}
Thus, we have
\begin{align}
    P_{\textrm{meet}} \geq \E{}{\sum_{k=2}^K \min(\widetilde{w}_k, \widetilde{v}_k)}.
\end{align}

By \Cref{ass:weightsbounded}, we have $w_1+S \leq K w_{\decp,\encp}^{\textrm{max}}$ (and similarly for $v_1+S$); thus we can further lower bound the probability of coupling,
\begin{align}\label{eq:boundonPmeet}
    P_{\textrm{meet}} & \geq \E{}{\sum_{k=2}^K \min\left(\frac{w_k}{w_1 + S}, \frac{v_k}{v_1 + S} \right)} \\
    & \geq \E{}{\frac{1}{K w_{\decp,\encp}^{\textrm{max}}} \sum_{k=2}^{K} w_k}
    = \frac{p_{\decp}(\obs)}{ w_{\decp,\encp}^{\textrm{max}}}-\frac{p_{\decp}(\obs)}{K w_{\decp,\encp}^{\textrm{max}}}.\nonumber
\end{align}
We have used above that $S=\sum_{k=2}^K w_k$ only depends on the $K-1$ proposals common to the two chains at any iteration and so $ \E{}{S}=\sum_{k=2}^K \E{}{w_k}$, where $\E{}{w_k}=\E{q_{\encp}}{w_{\decp,\encp}(\lat^{\star}_k)}=p_{\decp}(\obs)$. Contrary to \cite{jacob2019smoothing}, the lower bound we obtain on $P_{\text{meet}}$ is (as expected) increasing with $K$ instead of decreasing. From this lower bound, we can also deduce directly an upper bound on $\mathbb{E}[\tau]$:
\begin{equation}\label{eq:boundontau}
    \mathbb{E}[\tau]\leq (L-1)+\frac{1}{P_{\textrm{meet}}}
\end{equation}
where $1/P_{\textrm{meet}}$ is the expectation of a geometric random variable of success probability $P_{\textrm{meet}}$. By plugging the lower bound on the r.h.s. of \eqref{eq:boundonPmeet} in \eqref{eq:boundontau}, we obtain a decreasing upper bound on $\mathbb{E}[\tau]$ converging to $L-1+\frac{w_{\decp,\encp}^{\textrm{max}}}{p_{\decp}(\obs)}$ as $K\rightarrow \infty$.

\section{ADDITIONAL EXPERIMENTAL DETAILS}
\label{app:sec:experiments}

\subsection{Gradients of the \acrshort{PPCA} Model}
\label{app:subsec:ppca_grads}

\begin{figure}[t]
    \centering
    \includegraphics[width=0.8\columnwidth]{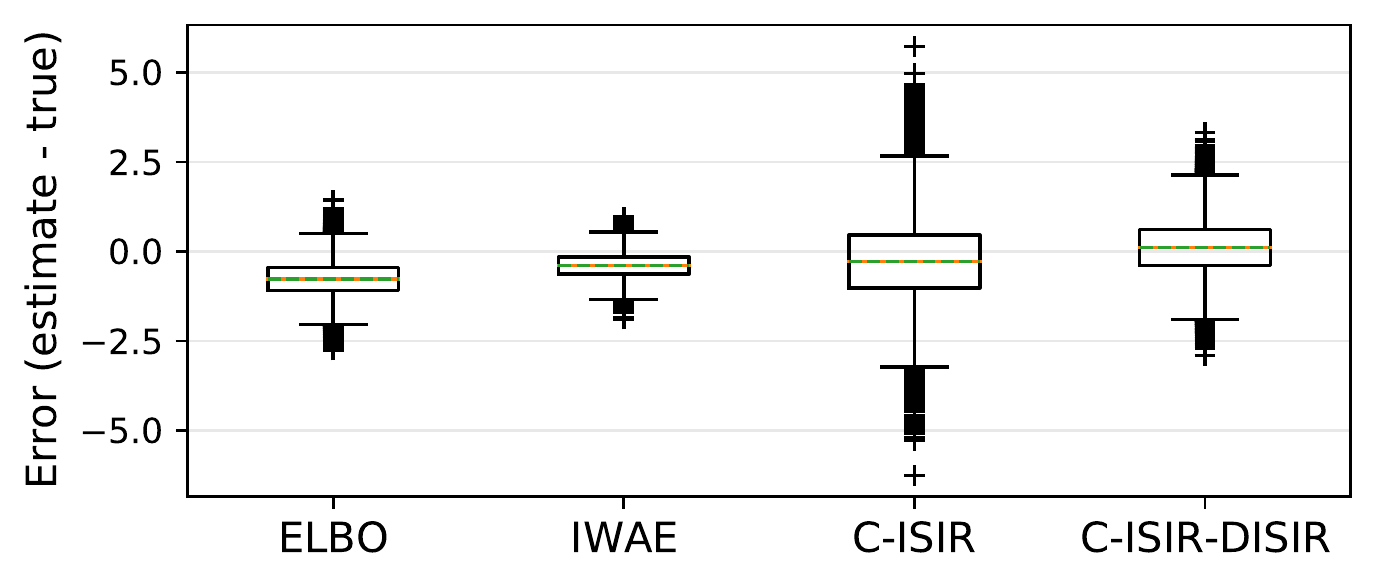}
    \vspace*{-8pt}
    \caption{Boxplot representation of the error of different estimators for the gradient w.r.t.\ one of the weights of the \acrshort{PPCA} model.%
    \label{fig:ppca_boxplots2}}
    \vspace*{-8pt}
\end{figure}

\Cref{fig:ppca_boxplots2} shows the errors when estimating the gradient w.r.t.\ a randomly chosen weight term of the \gls{PPCA} model; it looks qualitatively similar to \Cref{fig:ppca_boxplots}.

\begin{figure}[t]
    \centering
    \includegraphics[width=0.8\columnwidth]{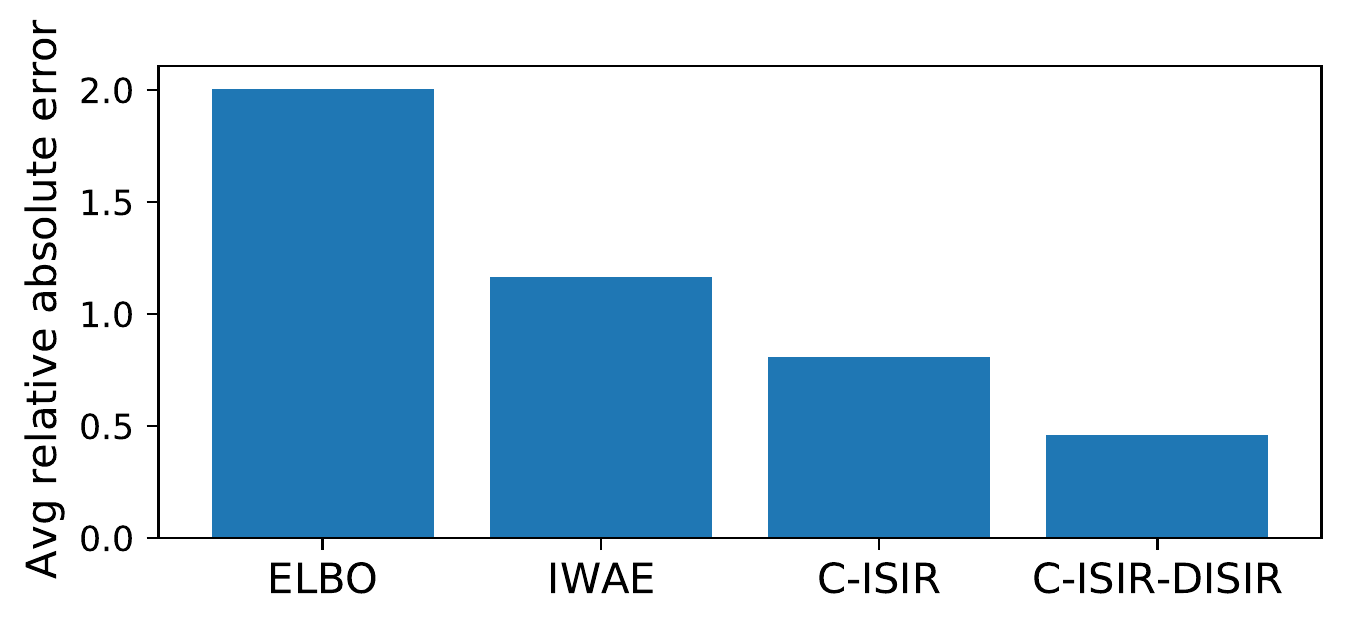}
    \vspace*{-8pt}
    \caption{Relative error (in absolute value) for the \acrshort{PPCA} model, averaged over all components of the gradient.%
    \label{fig:ppca_boxplots_all}}
    \vspace*{-8pt}
\end{figure}

\Cref{fig:ppca_boxplots_all} shows the relative error in absolute value, averaged across all the components of the gradient. C-\acrshort{ISIR}-\acrshort{DISIR} exhibits the smaller error.

\subsection{Experimental Setup for the \acrshort{VAE}}
\label{app:subsec:vae_details}

\parhead{Binarized MNIST.}
For binarized MNIST, the model is $p_{\decp}(\obs, \lat) = \Ncal(\lat; 0, I)p_{\decp}(\obs \g \lat)$, where $p_{\decp}(\obs \g \lat)$ is a product of Bernoulli distributions whose parameters are obtained as the output of a neural network with $2$ hidden layers of $200$ hidden units each and ReLU activations (the third layer is the output layer and has sigmoid activations). We set the distribution $q_{\encp}(\lat \g \obs)$ as a fully factorized Gaussian, and the encoder network has an analogous architecture (in this case, the output layer implements a linear transformation for the variational means and a softplus transformation for the standard deviations). The RMSProp learning rate is $5\times 10^{-4}$ and the batchsize is $100$.

\begin{figure}[t]
    \centering
    \includegraphics[width=0.45\textwidth]{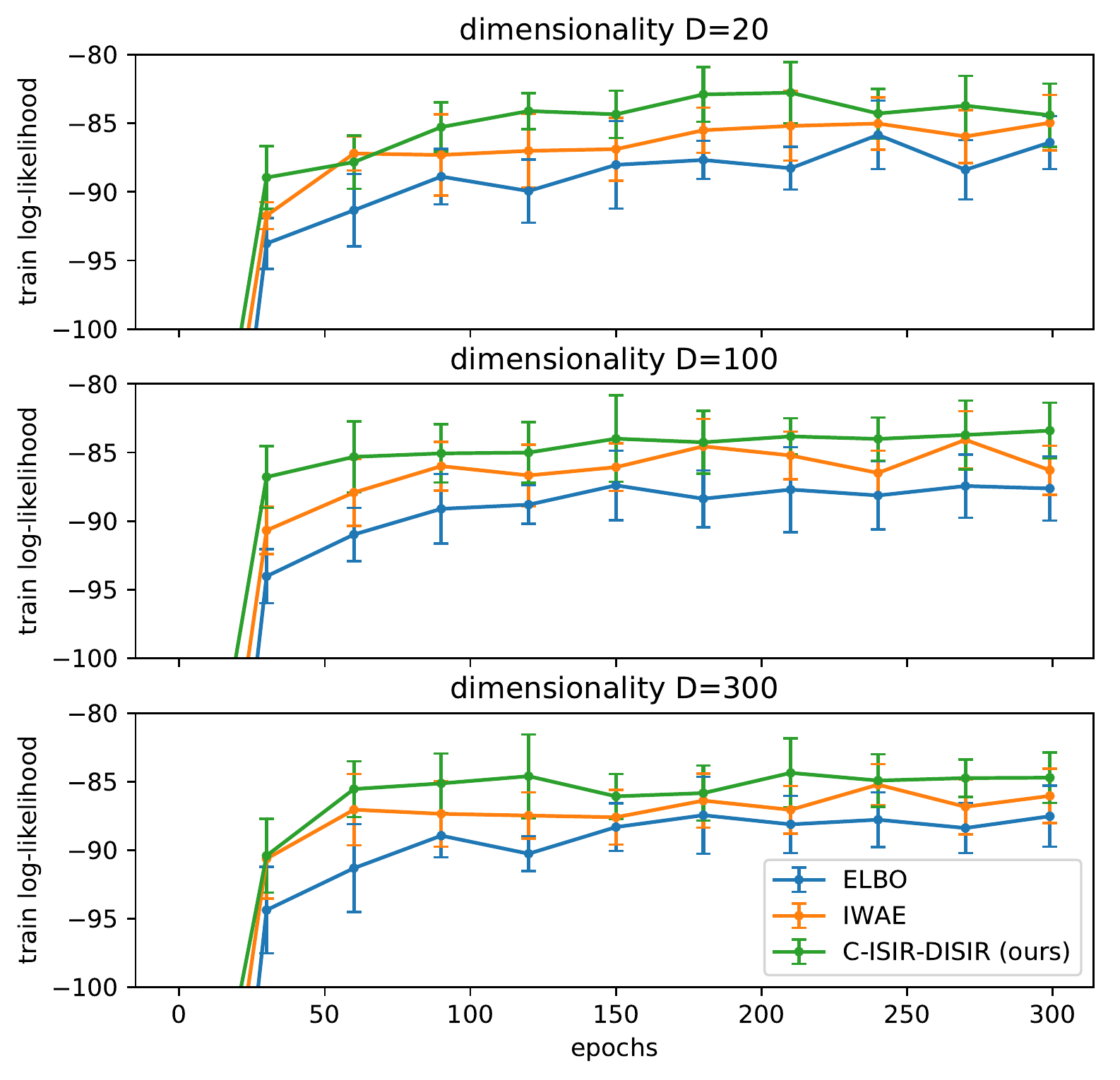}
    \caption{Train log-likelihood on a \acrshort{VAE} fitted on binarized MNIST. The estimator from \Cref{alg:couplingISIR} provides better performance for multiple values of the dimensionality $D$.%
    \label{fig:bmnist_loglik_all}}
\end{figure}

\parhead{Fashion-MNIST and CIFAR-10.}
We define the likelihood $p_{\decp}(\lat \g \obs)$ using a mixture of $10$ discretized logistic distributions \citep{salimans2017improving}.

For fashion-MNIST, we use the same encoder and decoder architecture as for binarized MNIST described above (except for the output layer of the decoder, which implements a linear transformation for the location parameters, a softplus transformation for the scale parameters, and a softmax transformation for the mixture weights).

For CIFAR-10, we use convolutional networks instead. The decoder consists of a fully connected layer with hidden size $16\times 16 \times 1$ and ReLu activations, followed by three convolutional layers with $200$, $50$, and $30$ channels (the filter size is $4\times 4$ with stride of $2$) and ReLu activations (except for the last layer). The encoder network has three convolutional layers with $64$, $128$, and $512$ channels, followed by a fully connected hidden layer with output size $128$ and by the output layer, which is the same as for fashion-MNIST.

The RMSProp learning rate is $10^{-4}$, and the batchsize is $100$ for fashion-MNIST and $50$ for CIFAR-10.

\subsection{Train Log-Likelihood on Binarized MNIST}
\label{app:subsec:train_log_lik}

\Cref{fig:bmnist_loglik_all} shows (an estimate of) the evolution of the train log-likelihood for the \acrshort{VAE} fitted on binarized MNIST for different values of the dimensionality $D$.

\begin{figure}[t]
    \centering
    \includegraphics[width=0.45\textwidth]{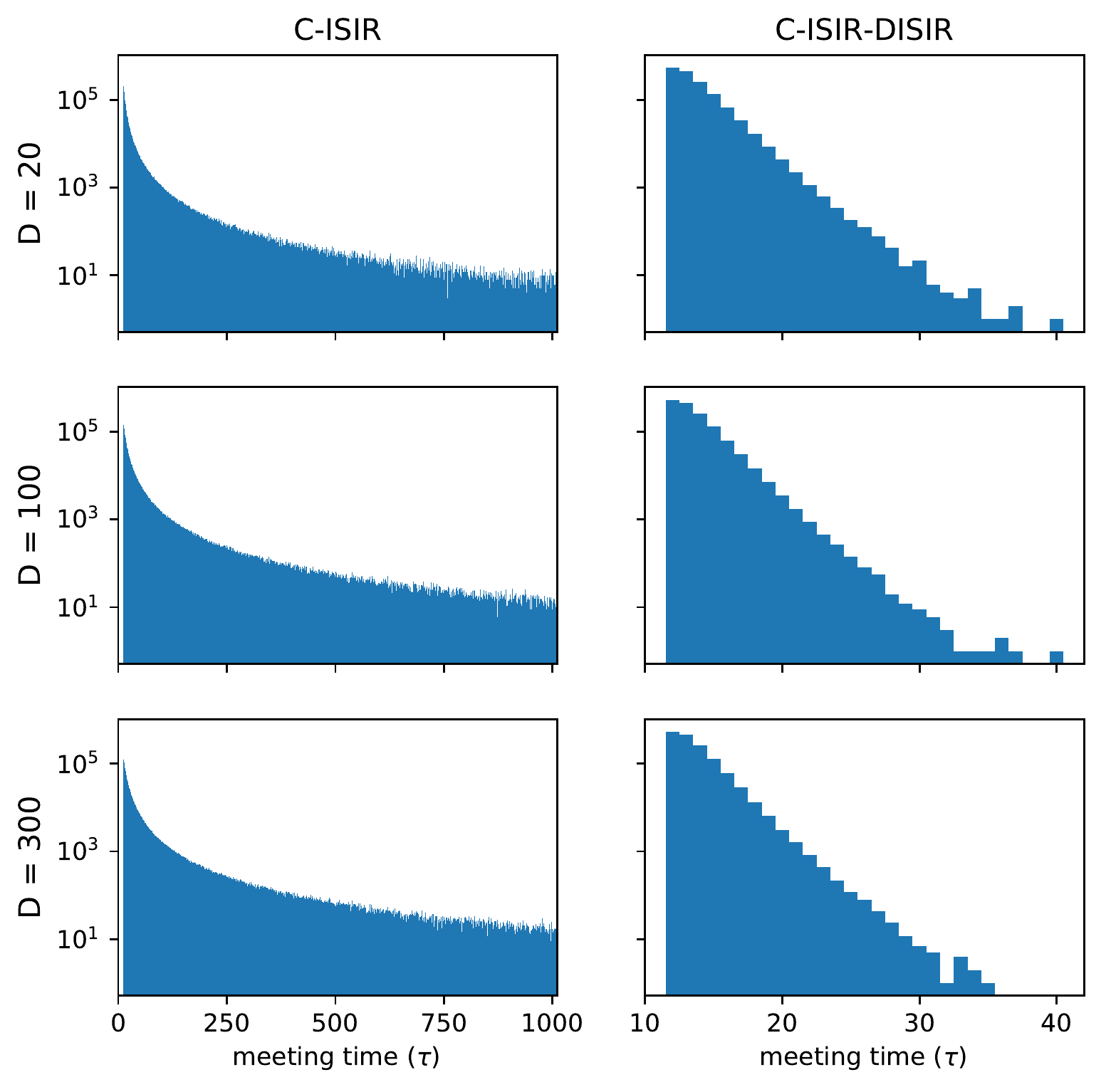}
    \vspace*{-5pt}
    \caption{Histograms of the meeting time for a \acrshort{VAE} fitted on binarized MNIST. The histograms corresponding to C-\acrshort{ISIR} have significantly heavier tails, which results in higher computational complexity of the overall estimator.%
    \label{fig:bmnist_histograms_tau_all}}
\end{figure}

\subsection{Histograms of the Meeting Time}
\label{app:subsec:hist_meeting_time}

\Cref{fig:bmnist_histograms_tau_all} shows the histograms of the meeting time for the experiment on binarized MNIST from \Cref{subsec:experiments_vae}. The meeting time behaves similarly across different values of the dimensionality $D$, although the histogram gets heavier-tailed for C-\acrshort{ISIR} when $D$ increases.

\end{document}